\documentclass{IEEEojcsys}
\usepackage[colorlinks,urlcolor=black,linkcolor=black,citecolor=black]{hyperref}

\usepackage{array}
\usepackage{graphicx}
\usepackage{amsmath,amssymb,amsfonts,amsthm}
\usepackage{algorithmic}
\usepackage{algorithm}
\usepackage{textcomp}
\usepackage{booktabs}
\usepackage{multirow}
\usepackage{url}
\usepackage{comment}
\usepackage[table,dvipsnames]{xcolor}

\newtheoremstyle{ieee_colon_style}
  {5pt} 
  {5pt} 
  {\normalfont} 
  {} 
  {\itshape} 
  {:} 
  {0.5em} 
  {\thmname{#1}\thmnumber{ #2}\thmnote{ (#3)}} 

\theoremstyle{ieee_colon_style}
\newtheorem{theorem}{Theorem}

\newtheorem{definition}[theorem]{Definition}
\newtheorem{problem}[theorem]{Problem}

\newtheorem{remark}[theorem]{Remark}

\jvol{00}
\jnum{XX}
\paper{25-0097}
\receiveddate{29 November 2025}
\accepteddate{29 April 2026}
\publisheddate{XX Month 2026}
\currentdate{XX Month 2026}
\pubyear{2026}
\doiinfo{OJCSYS.2026.3689712}

\begin{document}

%
\sptitle{Regular Paper}

\title{Robustness Certificates \\for Neural~Networks against Data~Poisoning and Evasion~Attacks}

\author{Sara Taheri\affilmark{1}, Mahalakshmi Sabanayagam\affilmark{2}, \\ Debarghya Ghoshdastidar\affilmark{2}, Majid Zamani\affilmark{1,3}}

\affil{LMU Munich, Munich, Germany (e-mail: sara.taheri@lmu.de)}
\affil{Technical University of Munich (TUM), Munich, Germany (e-mail: maha.sabanayagam@tum.de, ghoshdas@cit.tum.de)}
\affil{University of Colorado Boulder, Boulder, CO, USA (e-mail: majid.zamani@colorado.edu)}

\corresp{Corresponding Author: Sara Taheri (e-mail: sara.taheri@lmu.de)}

\authornote{This work is supported by the ConVeY DFG research training group and partially by NSF grants CNS-2039062 and CNS-2111688.}

\markboth{ROBUSTNESS CERTIFICATES FOR NEURAL NETWORKS}{TAHERI {\itshape ET AL}.}


\begin{abstract}
The increasing use of machine learning in safety-critical domains amplifies the risk of adversarial threats, especially  {data poisoning attacks} that corrupt training data to degrade performance or induce unsafe behavior. Most existing defenses lack formal guarantees or rely on restrictive assumptions about the model class, attack type, extent of poisoning, or point-wise certification, limiting their practical reliability. This paper introduces a principled formal robustness certification framework that models gradient-based training as a  {discrete-time dynamical system} (dt-DS) and formulates poisoning robustness as a formal safety verification problem. By adapting the concept of  {barrier certificates} (BCs) from control theory, we introduce sufficient conditions to certify a robust radius ensuring that the terminal model remains safe under worst-case ${\ell}_p$-norm based poisoning. 
To make this practical, we parameterize BCs as neural networks trained on finite sets of poisoned trajectories. We further derive  {probably approximately correct} (PAC) bounds by solving a  {scenario convex program} (SCP), which yields a confidence lower bound on the certified robustness radius generalizing beyond the training set.
Importantly, our framework also extends to certification against test-time attacks, making it the  {first} unified framework to provide formal guarantees in both training and test-time attack settings.
Experiments on MNIST, SVHN, CIFAR-10, and CIFAR-100 show that our approach certifies non-trivial perturbation budgets while being model-agnostic and requiring no prior knowledge of the attack or contamination level.
\end{abstract}


\maketitle

\section{INTRODUCTION}
\label{sec:intro}

The deployment of machine learning (ML) models in safety-critical domains, such as autonomous driving and medical diagnostics, increases the risk of adversarial threats, especially  {data poisoning attacks}. In such attacks, an adversary deliberately injects crafted perturbations into the  {training data set} to subvert the model's behavior, degrade performance, or violate safety requirements at test-time~\cite{biggio2012poisoning,shafahi2018poison,koh2017understanding}. In many real-world pipelines, training data is sourced and refreshed through a broader data supply chain (e.g., third-party datasets, labeling/annotation workflows, or periodic retraining on newly collected data), which can be influenced upstream within a bounded corruption budget \cite{goldblum2023dataset,carlini2024poisoning}. These attacks exploit the training pipeline, embedding backdoors or stealth vulnerabilities that can persist unnoticed and trigger failures in mission critical applications~\cite{carlini2024poisoning,schwarzschild2021toxic}. 

Although a variety of defenses have been proposed to mitigate data poisoning attacks, ranging from detecting and removing poisoned samples to modifying training strategies for robustness, these approaches are largely heuristic and remain vulnerable to sophisticated adaptive attacks \cite{goldblum2023dataset, koh2022stronger, shafahi2018poison, huang2020metapoison}. This highlights the need to develop  {formal robustness certificates} that guarantee that the predictions of a model remain unchanged by poisoning.

\begin{figure}
\centerline{\includegraphics[width=20.3pc]{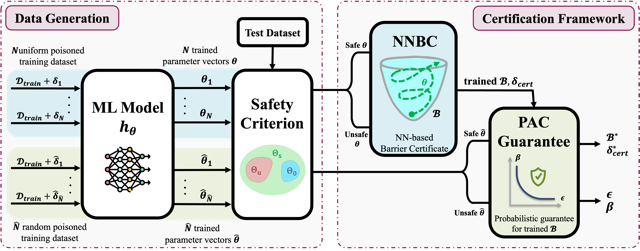}}
\caption{Overview of our proposed framework against train-time attacks. (Left) Data Generation: The model $h_\theta$ is trained on multiple poisoned datasets with varying perturbation levels to generate a set of parameter trajectories. The terminal parameters are labeled as safe or unsafe based on the test accuracy degradation. (Right) Certification: A Neural Network-based Barrier Certificate (NNBC) $\mathcal{B}$ is learned from these parameters. The validity of $\mathcal{B}$ is then rigorously verified via a PAC bound guarantee, providing a certified robust radius with a violation probability of at most $\epsilon$ and a confidence of at least $1\! -\! \beta$.}\label{fig:frame}
\vspace{-10pt}
\end{figure}

A small but growing line of work explores such robustness certification under fixed-threat models and a certain allowed corruption budget for poisoning. Notable techniques include randomized smoothing \cite{RAB}, model ensembling \cite{levine2021deep}, parameter-space interval bounds through convex relaxation \cite{sosni2024abstractgradient}, and combining kernels and linear programming approaches for large-width networks \cite{sabanayagam2025exact, gosch2025provable}. However, these methods face three major limitations:  
$(i)$ {Threat model and budget assumptions}: Most work assumes a fixed (un)bounded corruption budget, with {no} mechanism to compute the budget corresponding to a desired robustness level. Furthermore, it is assumed that the number of corrupted data points is known \cite{RAB};
$(ii)$  {Model specificity}: The approaches are limited to specific architectures in some cases, such as decision trees \cite{meyer2021certifying}, nearest neighbors \cite{KNN}, or graph neural networks \cite{sabanayagam2025exact,gosch2025provable}, and assume white-box access;  
$(iii)$  {Pointwise guarantees}: Most certification methods provide guarantees only for individual test points, failing to account for the global behavior of the model in the test data \cite{levine2021deep}.
Together, these limitations underscore a fundamental open question:  
\begin{quote} \emph{\small 
    ``Can we certify, for any ML model, an ${\ell}_p$-norm bounded poisoning budget such that the model’s performance degradation under poisoning is guaranteed to remain below a threshold $\alpha$?"
}\end{quote}
\normalfont

\noindent In this work, we answer this question positively by developing a framework inspired by  {control-theoretic safety verification} to certify ML models against data poisoning. We model gradient-based training as a discrete-time dynamical system (dt-DS), where the model parameters form the system state and the (potentially poisoned) training data act as the input to the system. 

Within this dynamical system view, we recast poisoning robustness as a formal safety verification problem and adapt  {barrier certificates (BCs)}~\cite{ames2016control,prajna2007framework} to certify a robust radius $\ell_p$ for a prescribed accuracy-degradation tolerance $\alpha$, ensuring that parameter trajectories remain within the safe set under worst-case poisoning. This enables principled worst-case guarantees without requiring knowledge of the specific ML architecture, the poisoning attack strategy, or the fraction of corrupted data, and provides a certificate for the  {entire} test dataset, not just point-wise test samples.

We note that it is challenging to explicitly construct the BCs for ML training processes due to the high dimensionality of the parameter space, lack of a closed-form training model, and the unknown nature of the poisoning attack model, thus rendering the exact system dynamics inaccessible. To address this, we adopt a {data-driven approach} that parameterizes BC as a neural network, producing a  {neural network-based BC (NNBC)}, similar to recent data-driven safety verification using BC for unknown systems~\cite{Mahathi,Clark,Abate}. 

Although NNBC is trained on a finite set of poisoned trajectories, we ensure that BC conditions hold more generally by reformulating verification as a  {scenario convex program} (SCP), allowing us to derive  {probably approximately correct (PAC)} bounds~\cite{campi2008exact,Abate}, providing a probabilistic guarantee. The PAC bound ensures, with some confidence, that the probability of violating the barrier conditions on unseen trajectories stays below a prescribed level. Figure~\ref{fig:frame} presents our proposed framework for train-time attacks. Importantly, the NNBC allows for certifying test-time corruptions as well, providing a unified approach to certify both train and test data poisoning. 
\subsection{CONTRIBUTION}
    \textbf{1.} We cast gradient-based ML training as a discrete-time dynamical system and reformulate robustness certification against train and test data perturbations as a formal  {safety verification problem} using {barrier certificates (BC)}. \\
    \textbf{2.} We introduce a neural network-based BC (NNBC) framework to overcome the intractability of the explicit BC design for high-dimensional and unknown poisoned training dynamics. NNBC is trained to obtain the certified robust radius, the largest admissible perturbation of the train or test data for which the degradation in test accuracy is provably at most a given threshold.\\
    \textbf{3.} We derive a  {probably approximately correct} (PAC) bound that provides a rigorous probabilistic guarantee for the trained NNBC and its associated certified robust radius. \\
    \textbf{4.} Our approach is model-agnostic and does not require prior knowledge of the ML architecture, the attack strategy, or the amount of data corrupted, thus broadly applicable. \\
    \textbf{5.} We demonstrate the effectiveness of our certification framework on various models and datasets, demonstrating its ability to quantify and formally certify safe perturbation budgets for training and test-time attacks in practice.

\subsection{RELATED WORK}\label{sec:related}
Although robustness certification against test-time adversarial attacks has been extensively studied, the literature on formal certificates for data poisoning remains significantly less developed. In particular, there has been limited progress on computing a certified $\ell_p$-norm poisoning radius that guarantees a desired model accuracy. Existing approaches to certify robustness against data poisoning can be broadly grouped into four categories. To clarify our positioning with respect to control-theoretic methods, we note upfront that the poisoning-specific certification works reviewed below are predominantly not control-theoretic (e.g., they are ensemble-, smoothing-, privacy-, or model/relaxation-based), whereas our framework is explicitly control-theoretic via barrier certificate-based verification of training dynamics.

 {(i) Ensemble-based methods.}
Ensemble-based certifications typically partition the training data set into multiple disjoint subsets and train the base classifiers {independently} on these subsets. A final ensemble classifier aggregates predictions (e.g., via majority voting or run-off elections), and robustness is certified by analyzing the minimum number of clean samples required to dominate poisoned samples in the ensemble decision rule \cite{levine2021deep,jia2021intrinsic,cohen2019certified,wang2022improved,rezaei2023runoff}. These methods generally assume independence across base models and often allow  {unbounded perturbation budgets} on poisoned samples, focusing instead on bounding the fraction of corrupted training points.

 {(ii) Randomized smoothing for poisoning.}
Randomized smoothing, originally developed as a test-time certification technique \cite{cohen2019certified}, has been adapted to data poisoning by injecting randomness into the training pipeline. These methods certify robustness by averaging model behavior over randomly perturbed training datasets, effectively assuming a  {fixed bounded perturbation} and providing guarantees on label or pattern corruptions. Existing work certifies robustness against label corruption \cite{rosenfeld2020certified}, specific backdoor patterns injected into subsets of training and test points \cite{RAB,wang2020certifying}, and joint feature and label-level corruptions \cite{zhang2022bagflip}. However, these approaches are typically tied to particular corruption models (e.g., fixed backdoor triggers) and do not directly yield a certified poisoning radius for general poisoning processes.

 {(iii) Differential-privacy-based methods.}
Another line of work uses theoretical connections between differential privacy and robustness to construct poisoning certificates. Here, differential privacy guarantees are used to bound the influence of individual training samples, thereby limiting the effect of poisoned points on the final model \cite{ma2019data,xie2023unraveling}. Although conceptually attractive, these certificates often inherit the conservatism of privacy guarantees and may require strong assumptions or substantial noise injection, which can degrade clean performance.

 {(iv) Model-specific certification methods.}
A further set of methods focuses on specific model families and assumes both  {bounded perturbation budgets} and a  {bounded number of poisoned samples}. For graph neural networks, \cite{gosch2025provable} employs the kernel-equivalence of neural networks through graph neural tangents \cite{sabanayagam2023analysis} to formulate mixed-integer linear programming-based certificates for $\ell_p$-norm feature corruptions, requiring explicit knowledge of corrupted training data and perturbation magnitudes. \cite{sabanayagam2025exact} extends this framework to handle label corruptions. \cite{sosni2024abstractgradient} proposes a gradient-based certification method for neural networks that uses convex relaxations and interval bounds over parameter trajectories under known corruption levels. However, these relaxations tend to become loose as training progresses, and the resulting certificates are tightly coupled to particular architectures and training setups, making generalization to broader model classes difficult. More broadly, such model-specific approaches rely on restrictive assumptions about the adversary’s behavior and detailed white-box information about the training process.


In contrast to these previous methods, our proposed framework provides a  {general-purpose} approach to certifying a robust poisoning radius based on the underlying training dynamics. By modeling training as a discrete-time dynamical system and leveraging BCs from control theory, we enable formal certification across both train-time and test-time poisoning settings. Crucially, our framework does not require model-specific architectural assumptions, detailed adversary knowledge, or specialized white-box access beyond standard gradient information, and it applies to a wide class of models trained with  gradient-based optimizers.

\section{PRELIMINARIES}
\label{sec:prelim}
All proofs are deferred to the Appendix.
\subsection{NOTATIONS}
We denote the sets of real, positive real, and negative real numbers by $\mathbb{R}$, $\mathbb{R}^+$, and $\mathbb{R}^-$, respectively. The absolute value of a scalar $x \in \mathbb{R}$ is denoted by $|x|$. The sets of positive integers and non-negative integers are denoted by $\mathbb{N}$ and $\mathbb{N}_0$, respectively.
The set $\mathbb{R}^d$ denotes the $d$-dimensional Euclidean space. We define $[r]$ as the set of the first $r$ natural numbers (i.e., $[r] := \{1, 2, \dots, r\}$). For any vector $x \in \mathbb{R}^d$, its Euclidean ($\ell_2$-norm) is denoted by $\|x\|_2$, and its infinity norm ($\ell_\infty$-norm) is denoted by $\|x\|_\infty := \max_i |x_i|$. The complement of a set $A \subseteq B$ within a universal set $B$ is denoted by $B\backslash A$. For any $a\in \mathbb{R}$, the ceiling function $\lceil a \rceil$ returns the smallest integer greater than or equal to a, and the Rectified Linear Unit (ReLU) activation is defined as $\operatorname{ReLU}(a) := \max\{0, a\}$.  
Finally, for any set $\Theta $, the indicator function $\mathbf{1}_\Theta (\theta)$ equals 1 if $\theta \in \Theta $ and 0 otherwise.

\subsection{MACHINE LEARNING SETUP FORMULATION}\label{subsec_setup_formulation}
We consider a clean training dataset 
$\mathcal{D}_{\mathrm{train}} \!=\! \{(u_i, y_i)\}_{i=1}^n \!\subseteq\! \mathbb{R}^m \!\times\! \mathcal{Y}$,
where $\mathcal{Y} \!:=\! \{1, \dots, k\}$ is a finite set of class labels. Each feature vector $u_i \!\in\! \mathbb{R}^m$ is paired with a label $y_i \!\in\! \mathcal{Y}$. A held-out test dataset is denoted by 
$\mathcal{D}_{\mathrm{test}} \!=\! \{(u_i', y_i')\}_{i=1}^{n'} \!\subseteq\! \mathbb{R}^m \!\times\! \mathcal{Y}$, which is used to evaluate the trained ML model.
Let $h_{\theta}\!:\! \mathbb{R}^m \!\rightarrow\! \mathbb{R}^k$ denote a parameterized machine learning (ML) model (e.g., a neural network) with parameter vector $\theta \!\in\! \mathbb{R}^d$. This function maps inputs to a vector of continuous-valued outputs (e.g., logits). The model is trained by a gradient-based optimization rule to minimize the empirical loss 
   $ \mathcal{L}(h_\theta, \mathcal{D}_{\mathrm{train}}) \!:=\! \frac{1}{n} \sum_{i=1}^n \ell\big(h_\theta(u_i), y_i\big)$,
where $\ell\!:\!\mathcal{Y}\!\times\!\mathcal{Y} \!\to\! \mathbb{R}^+$ is a non-negative differentiable surrogate loss function (e.g., cross-entropy loss).
Training is performed for a terminal training horizon $t_\infty$ (determined by a pre-defined number of epochs or upon the minimization of the loss function).
Note that the model $ h_\theta $ outputs a continuous $ k $-dimensional vector rather than a discrete label in $ \mathcal{Y} $. Each of the $ k $ components of this vector corresponds to the score of the model for a particular class. Such a continuous output is essential for gradient-based training, as the loss function must be differentiable with respect to these outputs to enable gradient computation. The final discrete class prediction is obtained only when required (e.g., during evaluation) by selecting the index of the maximum score within this vector.

In ideal settings, ML models are trained on clean training datasets and evaluated under the assumption that the test dataset is uncorrupted.
However, in practice, the data used to train or evaluate an ML model $h_{\theta}$ may be adversarially perturbed, resulting in a poor performance. Such poisoning attacks can target input features, labels, or both, and may occur during either the training or testing phases of the ML pipeline. In this work, we focus on input-space poisoning, where perturbations affect the training or test data features. Our certification framework provides formal guarantees of the maximum allowable perturbation magnitude, measured in the $\ell_p$ norm. The following definitions formalize this poisoning threat model.

\begin{definition}[Train-Time Poisoning attack]\label{def:poisoning}
Let $\mathcal{D}_{\mathrm{train}}=\{(u_i, y_i)\}_{i=1}^n$ be the clean training set. A poisoning attack is modeled as an adversary $\mathcal{A}$ that perturbs an unknown fraction $\rho\in[0,1]$ of the training samples to corrupt the training process, so that the model learned from the poisoned data performs worse on downstream evaluation data than the model trained on $\mathcal{D}_{\mathrm{train}}$.
Formally, the adversary modifies $r:=\lceil \rho n \rceil$ inputs, yielding
\begin{align}
\mathcal{D}_{\mathrm{train}}^\Delta := \big\{(u_i+\delta_i, y_i)\big\}_{i=1}^r \cup \big\{(u_i, y_i)\big\}_{i=r+1}^n,
\end{align}
{where $\Delta := [\delta_1 \dots \delta_r] \in \mathbb{R}^{r\times m}$ is the poisoning perturbation matrix, and
\begin{align}
\Lambda := \{\Delta \in \mathbb{R}^{r\times m} \mid \|\Delta\|_p := \max_{i\in[r]} \|\delta_i\|_p \le \delta\},
\end{align}
is the set of admissible train-time perturbations for some $\delta \ge 0$.}
If $\delta=0$ or $\rho=0$, then $\mathcal{D}_{\mathrm{train}}^\Delta \equiv \mathcal{D}_{\mathrm{train}}$.
\end{definition}

\begin{definition}[Test-Time Evasion attack]\label{def:test_evasion}
Let $\mathcal{D}_{\mathrm{test}}=\{(u_i', y_i')\}_{i=1}^{n'}$ be the clean test set. An evasion attack is modeled as an adversary $\mathcal{A}'$ that perturbs an unknown fraction $\rho'\in[0,1]$ of the test samples to induce prediction errors or increase the loss on perturbed inputs, while keeping the trained model parameters fixed.
Formally, the adversary modifies $r':=\lceil \rho' n' \rceil$ inputs, yielding
\begin{align}
\mathcal{D}_{\mathrm{test}}^{\Delta'} := \big\{(u_i'+\delta_i', y_i')\big\}_{i=1}^{r'} \cup \big\{(u_i', y_i')\big\}_{i=r'+1}^{n'},
\end{align}
{where $\Delta' := [\delta_1' \dots \delta_{r'}'] \in \mathbb{R}^{r'\times m}$ is the evasion perturbation matrix, and
\begin{align}
\Lambda' := \{\Delta' \in \mathbb{R}^{r'\times m} \mid \|\Delta'\|_p := \max_{i\in[r']} \|\delta_i'\|_p \le \delta'\},
\end{align}
is the set of admissible test-time perturbations for some $\delta' \ge 0$.}
If $\delta'=0$ or $\rho'=0$, then $\mathcal{D}_{\mathrm{test}}^{\Delta'} \equiv \mathcal{D}_{\mathrm{test}}$.
\end{definition}
Note that, for ease of exposition, we assume that the first $r$ elements of the training dataset and the first $r'$ elements of the test dataset are poisoned. However, our certification framework is permutation-invariant and does not require knowledge of the indices of corrupted data. 

Although training minimizes $\mathcal{L}$, the generalization performance of the model trained on a (possibly) poisoned dataset is evaluated on a (possibly) perturbed test dataset:
\begin{align}\label{eq:test-accuracy-stoch}
g(\mbox{\small$\theta;\mathcal{D}_{\mathrm{test}}^{\Delta'}$})
\!:=\! \frac{1}{n'} \sum_{i=1}^{n'} \mathbf{1}_{\left\{ \arg\max_{j \in [k]} (h_\theta(u_i'+\delta'_i))_j = y_i' \right\}},
\end{align}
which computes the fraction of test samples correctly classified by $h_\theta$. Each $\delta_i' \in \mathbb{R}^m$ is the perturbation applied to the $i$-th input in $\mathcal{D}_{\mathrm{test}}^{\Delta'}$, where $\delta_i' = \mathbf{0}$ for all unperturbed samples ($i > r'$), as in Definition~\ref{def:test_evasion}.
 The performance of the trained model is assessed using the final test accuracy denoted by $g{\mbox{\small$(\theta(t_\infty), \mathcal{D}_{\mathrm{test}}^{\Delta'})$}}$, where $\theta(t_{\infty})$ represents the model parameter vector at the terminal training index and $t_{\infty}$ refers to the endpoint of a  {finite} training trajectory used for certification (e.g., a prescribed number of training epochs (iterations), or the first iteration satisfying a stopping criterion such as loss convergence), and should not be interpreted as an asymptotic infinite-time limit.

\subsection{PROBLEM FORMULATION}\label{subsec_problem_formulation}
As discussed earlier, the ML lifecycle is vulnerable to adversarial interference at both training and test time. During training, data poisoning can distort the optimization trajectory and drive the parameters into suboptimal or unsafe regions of the parameter space $\Theta$, while at inference, evasion attacks exploit model sensitivities to bypass decision boundaries and degrade generalization and reliability. These vulnerabilities motivate robustness certification frameworks that provide formal guarantees on model behavior under adversarial perturbations. To quantify such resilience, the following definition formalizes the notion of model degradation, which we adopt as the primary robustness criterion for a trained ML model.

\begin{definition}[Model Degradation]
\label{def:deg}
Let $h_{\theta}$ be an ML model with parameters $\theta \!\in\! \mathbb{R}^d$, trained on a (possibly) poisoned dataset $\mathcal{D}_{\mathrm{train}}^{\Delta}$ and evaluated on a (possibly) perturbed test dataset $\mathcal{D}_{\mathrm{test}}^{\Delta'}$, using the performance evaluation function $g$ as in~\eqref{eq:test-accuracy-stoch}. To quantify the degradation in performance under adversarial attacks, we define the model degradation of $h_{\theta}$ at iteration $t'$ as the drop in test accuracy relative to the clean-data baseline:
\begin{align}\label{eq:deg}
&\mathcal{G}\!\left(\mbox{\small$\theta(t')$}\right):=g_\mathrm{c}\!\left(\mbox{\small$\theta(t'),\mathcal{D}_{\mathrm{test}}$}\right)
-
g_\mathrm{p}\!\left(\mbox{\small$\theta(t'),\mathcal{D}_{\mathrm{test}}^{\Delta'}$}\right).
\end{align}
where $g_{\mathrm{c}}(\mbox{\small$\theta(t'), \mathcal{D}_{\mathrm{test}}$})$ denotes the test accuracy of the model trained and evaluated on clean datasets, and $g_{\mathrm{p}}(\mbox{\small$\theta(t'), \mathcal{D}_{\mathrm{test}}^{\Delta'}$})$ denotes the test accuracy of the model trained on the (possibly) poisoned dataset $\mathcal{D}_{\mathrm{train}}^{\Delta}$ and evaluated on the (possibly) perturbed dataset $\mathcal{D}_{\mathrm{test}}^{\Delta'}$.
\end{definition}
\vspace{-0.1cm}
\begin{remark}
The quantity $g_{\mathrm{c}}$ is treated as a generic performance benchmark. 
In ideal settings with access to clean data, $g_{\mathrm{c}}(\mbox{\small$\theta(t'),\mathcal{D}_{\mathrm{test}}$})$ coincides with the clean-data test accuracy. 
More generally, when clean datasets are unavailable or when a specification-driven baseline is preferred, $g_{\mathrm{c}}$ can be set to a prescribed target level. 
Accordingly, the attacked performance $g_{\mathrm{p}}(\mbox{\small$\theta(t'),\mathcal{D}_{\mathrm{test}}^{\Delta'}$})$ captures the evaluation of the (possibly) poisoned training pipeline on a (possibly) perturbed test set (including the special cases $\Delta=\mathbf{0}$ or $\Delta'=\mathbf{0}$). 
This enables certification relative to a required satisfaction level, independently of the empirical clean accuracy.
\end{remark}

It is important to remark that, although Definition~\ref{def:deg} is stated for arbitrary training iterations, our focus is on certifying the robustness of the terminal model $h_\theta$ after $t_\infty$ iterations. In particular, we seek to determine a certified robust radius under $\ell_p$-norm perturbations such that the resulting model degradation remains below a prescribed threshold. This objective is formalized in the following problem definition.

\begin{problem}[Certified Robust Radius]
\label{prob:cert-robustness}
Let $h_{\theta}$ be a parameterized ML model trained on a (potentially) poisoned training set $\mathcal{D}_{\mathrm{train}}^{\Delta}$ and evaluated on a (potentially) poisoned test set $\mathcal{D}_{\mathrm{test}}^{\Delta'}$.  
Given a threshold $\alpha \!\in\![0,1]$, the objective is to determine the largest poisoning radius $\delta_{\mathrm{cert}}$ for training-time (resp. $\delta'_{\mathrm{cert}}$ for test-time), such that, for all perturbations $\Delta$ (resp. $\Delta'$) satisfying $\|\Delta\|_p \leq \delta_{\mathrm{cert}}$ (resp. $\|\Delta'\|_p \leq \delta'_{\mathrm{cert}}$), the degradation of the trained model remains within $\alpha$.
\end{problem}

In the subsequent section, we derive the theoretical framework and certification methodology to address Problem~\ref{prob:cert-robustness}.

\subsection{METHODOLOGY}
While empirical evaluations provide an estimate of robustness against specific attack algorithms, they fail to yield formal guarantees that hold across all trajectories within the admissible perturbation sets. To bridge this critical gap and rigorously address Problem~\ref{prob:cert-robustness}, we introduce a formal robustness certification framework. 
We adopt a systems-theoretic perspective, modeling the iterative training process as a discrete-time dynamical system (dt-DS). This abstraction allows us to perform a principled reachability analysis, tracking how adversarial perturbations propagate through the model parameter evolution. We formalize this representation in the following definition.

\begin{definition}[Dynamical System]
\label{def:system-dynamics}
Let $h_{\theta}$ be the ML model defined in Section~\ref{subsec_setup_formulation}. The training process is modeled as a discrete-time dynamical system (dt-DS), denoted by a tuple  
\begin{align}
\mathfrak{S} = (\Theta, \Theta_0, \mathcal{D}_{\mathrm{train}}^{\Delta}, f),
\end{align}
where  $ \Theta  \subseteq \mathbb{R}^d $ is the set of model parameters, $\Theta _0\subseteq \Theta $ is the set of initial model parameters, $\mathcal{D}_{\mathrm{train}}^{\Delta}$ is the (potentially poisoned) training dataset, and
$ f \!: \mathbb{R}^d \times \mathbb{R}^m \times \mathcal{Y}\to \mathbb{R}^d $
is the parameter update map governed by the optimization algorithm.
The system $\mathfrak{S}$ evolves according to:
\begin{align}\label{eq:dtDS}
\theta(t+1) = f\big(\theta(t),\mathcal{D}_{\mathrm{train}}^{\Delta}(t)\big), \quad \forall t \!\in\! \mathbb{N}_0,  \forall \theta(0)\!\in\! \Theta_0,
\end{align}
where $\theta(t) \in \Theta$ is the model state at iteration $t$ and $\mathcal{D}_{\mathrm{train}}^{\Delta}(t) \subseteq \mathcal{D}_{\mathrm{train}}^{\Delta}$ is the mini-batch input.
\end{definition}

For simplicity, we write the update map as
$f{\mbox{\small$(\theta,\Delta)$}}\!:=\!f{\mbox{\small$(\theta,\mathcal{D}_{\mathrm{train}}^\Delta)$}}$,
where $\Delta$ encodes the perturbation matrix applied to the training inputs. Based on this dynamical abstraction, we recast the robustness certification of $h_\theta$ as a formal safety verification problem. In particular, robustness is defined as a safety property: the ML model is deemed robust if the state trajectory of $\mathfrak{S}$ remains invariant within a safe set. The following definition formalizes the corresponding safety specification for the dynamical system $\mathfrak{S}$.

\begin{definition}[Safety Specification]
\label{def:safety_specific}
Consider a dt-DS $\mathfrak{S}$ describing the training process of an ML model $h_\theta$ 
as in Definition~\ref{def:system-dynamics}, with the test accuracy degradation function $\mathcal{G}$ as defined in~\eqref{eq:deg}. 
Given a threshold $\alpha \in [0, 1]$, we define the safe and unsafe parameter sets, as follows:
\begin{align}\label{eq:SafeUnsafeSet}
\Theta_s^{\Delta'} \!:=\! \left\{ \theta \!\in\! \mathbb{R}^d \;\middle|\; \mathcal{G}(\mbox{\small$\theta$}) \!\leq\! \alpha \,\right\}, 
\quad
\Theta_u^{\Delta'} \!:=\! \Theta \backslash \Theta_s^{\Delta'}.
\end{align}
\end{definition}

\begin{remark}
Definition~\ref{def:safety_specific} introduces safety in a state-space form by partitioning the parameter space into safe and unsafe sets with respect to the degradation threshold $\alpha$. This formulation is stated at the level of parameters $\theta$ (rather than only at a specific iteration) to provide a general notion of safety for the training dynamics. Nevertheless, the primary robustness objective in this work is terminal-time performance, i.e., ensuring that the final trained parameters $\theta(t_\infty)$ satisfy the safety condition.
\end{remark}

To formally verify the safety specification in Definition~\ref{def:safety_specific} for the system $\mathfrak{S}$, we then adopt a barrier certificate (BC) formulation, inspired by \cite{prajna2004}, as the foundation of our robustness certification framework.

\begin{definition}[Barrier Certificate]\label{def:barrier_function}
Let $h_{\theta}$ be an ML model as in Section~\ref{subsec_setup_formulation} with its associated dt-DS $\mathfrak{S}$ as defined in Definition~\ref{def:system-dynamics} and the corresponding safety specification 
as in Definition~\ref{def:safety_specific}. A function $\mathcal{B}: \mathbb{R}^d \!\to \!\mathbb{R}$ is called a barrier certificate (BC) for $\mathfrak{S}$ with respect to the initial set $\Theta_0\subseteq\Theta$, the safe set $\Theta_s^{\Delta'}\subseteq\Theta$, and the unsafe set $\Theta_u^{\Delta'}\subseteq\Theta$, if there exists a certified train-time robust radius $\delta_{\mathrm{cert}}$ (resp. test-time radius $\delta_{\mathrm{cert}}'$) such that the following conditions hold for all perturbations satisfying $\|\Delta\|_p \leq \delta_{\mathrm{cert}}$ (resp. $\|\Delta'\|_p \leq \delta_{\mathrm{cert}}'$):
\begin{align}
\label{B0}
&\mathcal{B}{\mbox{\small$(\theta)$}} \leq 0, &&\quad\forall\, \theta \in \Theta _0,
\\
\label{Bu}
&\mathcal{B}{\mbox{\small$(\theta)$}} > 0, &&\quad\forall\, \theta \in \Theta _u^{\Delta'},
\\
\label{Bs}
&\mathcal{B}(f{\mbox{\small$(\theta,\Delta)$}}) \leq 0, &&\quad\forall\, \theta \in \Theta , ~ \text{s.t.} ~ \mathcal{B}{\mbox{\small$(\theta)$}} \leq 0.
\end{align}
\end{definition}

\begin{remark}\label{rem:BC}
Conditions~\eqref{B0} and~\eqref{Bu} are  {requirements on the certificate} $\mathcal{B}$ with respect to the prescribed initialization set $\Theta_0$ and unsafe set $\Theta_u^{\Delta'}$. In particular, $\Theta_0$ is induced by the chosen initialization rule (e.g., random initialization within a specified distribution/support), and the certificate is constructed so that $\mathcal{B}(\theta)\le 0$ holds for all admissible initializations for all $\theta\in\Theta_0$. Note that starting from a safe initialization does not by itself guarantee robustness of the terminal model: training is necessary to attain the desired functionality, and the parameter trajectory may enter $\Theta_u^{\Delta'}$ as optimization proceeds. For this reason, we enforce the forward-invariance condition~\eqref{Bs} to ensure that the barrier sublevel set remains invariant under the poisoned training dynamics, which in turn guarantees safety of the final trained parameters.
In addition, the forward condition used in \eqref{Bs} is a drift/separation-style discrete-time barrier condition and is not intended to impose a multiplicative contraction. This is deliberate, since contraction-type conditions can be overly restrictive for the data-driven training dynamics considered in our robustness certification setting. 
\end{remark}

\begin{remark}
Train-time poisoning alters the dynamics $f{\mbox{\small$(\theta,\Delta)$}}$ and affects the reachability condition~\eqref{Bs}, while test-time poisoning modifies only $\mathcal{D}_{\mathrm{test}}^{\Delta'}$, influencing the output of the safety function $g{\mbox{\small$(\theta, \mathcal{D}_{\mathrm{test}}^{\Delta'})$}}$. Additional factors, such as model architecture, choice of optimizer, attack strength, and strategies also affect these trajectories. Nevertheless, because BC focuses only on parameter trajectories, it remains agnostic to all of these sources of variability, which is a significant advantage of our framework.
\end{remark}

\begin{theorem}[Certified Robust Radius]\label{thm:safety-guarantee}
Let $h_{\theta}$ be an ML model, trained on a (potentially) poisoned dataset $\mathcal{D}_{\mathrm{train}}^\Delta$ and evaluated on a (potentially) poisoned dataset $\mathcal{D}_{\mathrm{test}}^{\Delta'}$, as in Section~\ref{subsec_setup_formulation}. Given constant $\alpha\in[0,1]$, consider a dt-DS $\mathfrak{S} \!=\! (\Theta , \Theta _0,\mathcal{D}_{\mathrm{train}}^{\Delta}, f)$ as in Definition~\ref{def:system-dynamics}, modeling the training dynamics of $h_{\theta}$, and let $\Theta _s^{\Delta'} \!\subseteq\! \Theta $ and $\Theta _u^{\Delta'} \!\subseteq \!\Theta $ denote the corresponding safe and unsafe sets of terminal parameters, as introduced in Definition~\ref{def:safety_specific}. Suppose there exists a BC $\mathcal{B}$ for $\mathfrak{S}$ satisfying the conditions in Definition~\ref{def:barrier_function} for all perturbations $\|\Delta\|_p \!\leq \! \delta_\mathrm{cert}$ (resp. $\|\Delta'\|_p \!\leq \!\delta_\mathrm{cert}'$). 
Then, $\delta_\mathrm{cert}$ (resp. $\delta_\mathrm{cert}'$) serves as a certified train-time (resp. test-time) robust radius, ensuring that, under worst-case perturbations, the accuracy degradation is at most~$\alpha$.
\end{theorem}

Despite the theoretical existence of this certificate, constructing a BC $\mathcal{B}$ for the system $\mathfrak{S}$ is intractable due to the high dimensionality of the model parameters and the lack of an explicit mathematical model of $\mathfrak{S}$. To overcome these challenges, we propose a data-driven framework that learns $\mathcal{B}$ directly from sampled data, thereby bypassing analytical intractability.
Furthermore, since our approach is data-driven, we subsequently introduce a formal validation stage that provides guarantees on the reliability of the learned BC.

\section{DATA-DRIVEN ROBUSTNESS CERTIFICATION}
\label{sec:nn_barrier}
In this section, we present our data-driven approach to certify
the robustness of an ML model $h_{\theta}$ by constructing a neural
network-based barrier certificate (NNBC) that satisfies the conditions in Definition~\ref{def:barrier_function}.

\subsection{NNBC STRUCTURE}
Given a dynamical system $\mathfrak{S}$ as in Definition~\ref{def:system-dynamics}, we define an NNBC
$\mathcal{B}_{\varphi} : \mathbb{R}^{d} \to \mathbb{R}$ parameterized by $\varphi$.
The input to the network is the state vector $\theta \in \mathbb{R}^{d}$,
and the output is a scalar barrier value $\mathcal{B}_{\varphi}(\theta)$.
The network uses ReLU activations in the hidden layers to
ensure a piecewise affine, locally Lipschitz structure, and an
identity activation in the output layer. In addition, the depth and width
of $\mathcal{B}_{\varphi}(\theta)$ are tunable hyperparameters.

\subsection{DATA GENERATION}\label{subsec:data_gen}
To train an NNBC $\mathcal{B}_\varphi$, we generate a dataset by training $h_{\theta}$ under varying levels of data poisoning across $N$ uniformly spaced budgets. Specifically, we define two grids $
    \delta_{\mathrm{grid}}
    \!= \!\langle\delta_1,\dots,\delta_N\rangle,$ and $\delta'_{\mathrm{grid}}
    \!=\! \langle\delta_1',\dots,\delta_N'\rangle$,
where each point in $\delta_{\mathrm{grid}}$ and $\delta'_{\mathrm{grid}}$, represent train- and test-time poisoning levels, respectively.  Depending on the certification type (train- or test-time), one grid is fixed to zero. Then, for each $i \in [N]$, we randomly initialize $h_{\theta}$ at ${\theta}_i{\mbox{\small$(0)$}}\! \in \!\Theta_0$ and train it on $\mathcal{D}_{\mathrm{train}}^{\Delta}$ with $\|\Delta\|_p \!=\! \delta_i$, yielding the terminal state ${\theta}_i{\mbox{\small$(t_\infty)$}}$. Next, following Definition~\ref{def:deg}, each trained model is evaluated on the dataset $\mathcal{D}_{\mathrm{test}}^{\Delta'}$, where $\|\Delta'\|_p \!= \!\delta'_i$, and classified as safe or unsafe based on the safety criterion $\mathcal{G}(\mbox{\small$\theta_i(t_\infty)$})$ compared to a threshold $\alpha\! \in\! [0,1]$.
Thus, the resulting datasets are:
\begin{align}\label{I}
    \mathcal{I} &\!=\! \big\{ \theta_i(0) \!\mid\! \theta_i(0) \!\in\! \Theta_0 \big\},  \\
    \mathcal{S} &\!=\! \big\{ \theta_i(t_\infty) \!\mid\! \mathcal{G}(\mbox{\small$\theta_i(t_\infty)$}) \!\le\! \alpha \big\}, \\ \label{U}
    \mathcal{U} &\!=\! \big\{ \theta_i(t_\infty) \!\mid\!\mathcal{G}(\mbox{\small$\theta_i(t_\infty)$}) \!>\! \alpha \big\},
\end{align}
where $\mathcal{I} \subseteq \Theta_0$, $\mathcal{S} \subseteq \Theta_s^{\Delta'}$, and $\mathcal{U} \subseteq \Theta_u^{\Delta'}$
denote finite sampled parameter sets drawn from the initial, safe, and unsafe sets, respectively.
We define their union as $\vartheta := \mathcal{I} \cup \mathcal{S} \cup \mathcal{U} \subseteq \Theta$. Note that, depending on the chosen threshold $\alpha$, some sampled sets may be empty; we refer interested readers to the Appendix for more details.

Based on the generated dataset, we compute the empirical train-time robust radius $\delta_{\mathrm{emp}}$ (resp. test-time radius $\delta_{\mathrm{emp}}'$) as the largest perturbation bound such that, for all $\Vert\Delta\Vert_p \le \delta_{\mathrm{emp}}$ (resp. $\Vert\Delta'\Vert_p \le \delta_{\mathrm{emp}}'$), the terminal parameters $\theta\mbox{\small$(t_\infty)$}$ satisfy $\mathcal{G}(\mbox{\small$\theta(t_\infty)$})\!\le\! \alpha$ and thus remain in the safe set $\Theta_s^{\Delta'}$. These empirical radii act as optimistic upper bounds on the true robustness, since they are obtained from finite sampling rather than worst-case verification. Thus, any valid certified radius must satisfy the necessary condition $\delta_{\mathrm{cert}} \!\le\! \delta_{\mathrm{emp}}$.

\subsection{LOSS}
\label{subsec:loss}
To synthesize an NNBC $\mathcal{B}_{\varphi}$ that satisfies the conditions in Definition~\ref{def:barrier_function} on the datasets generated in Section~\ref{subsec:data_gen} for a given candidate perturbation radius $\delta_{\mathrm{cand}} \leq \delta_{\mathrm{emp}}$ (resp. $\delta'_{\mathrm{cand}} \leq \delta'_{\mathrm{emp}}$), we introduce a composite loss $\mathcal{L}$, computed on trajectories with perturbations bounded by $\lVert \Delta_i \rVert_p \leq \delta_{\mathrm{cand}}$ (resp. $\lVert \Delta'_i \rVert_p \leq \delta'_{\mathrm{cand}}$) and $i \!\in\! [N]$:
\begin{align} \nonumber 
  \mathcal{L}(\varphi)
=&~c_{\mathcal{I}}\!\sum_{{\theta}_i\in \mathcal{I}}
\!\mathrm{ReLU}\bigl(\mathcal{B}_{\varphi}{\mbox{\small$(\theta_i)$}}\bigr)+c_{\mathcal{U}}\!\sum_{{\theta}_i\in \mathcal{U}}
\!\mathrm{ReLU}\bigl(-\mathcal{B}_{\varphi}{\mbox{\small$(\theta_i)$}}\bigr) \\\label{Loss_NNBC} 
&+c_{\mathcal{Z}}\!\sum_{{\theta}_i\in \mathcal{Z}}
\!\mathrm{ReLU}\bigl(\mathcal{B}_{\varphi}{\mbox{\small$\bigl(f{\mbox{\small$(\theta_i,\Delta_i)$}}$}}\bigr).
\end{align}
 The scalars $c_{\mathcal{I}}, c_{\mathcal{U}}, c_{\mathcal{Z}} > 0$ weight the respective conditions, and the set $\mathcal{Z} \!\subseteq\! \vartheta$, is defined by
$\mathcal{Z} \!=\! \{{\theta}_i\!\mid\! {\theta}_i\!\in\!\vartheta, ~ \mathcal{B}_{\varphi}({\theta}_i)\!\leq\!0\}$. During training, the optimization process aims to minimize~$\mathcal{L}$ as closely as possible to zero. A zero loss ($\mathcal{L}(\varphi)=0$) corresponds to an  {idealized} empirical feasibility event: the learned NNBC $\mathcal{B}_{\varphi}$ (together with the associated candidate robust radius $\delta_{\mathrm{cert}}$ (resp.\ $\delta'_{\mathrm{cert}}$)) satisfies the constraints~\eqref{B0}--\eqref{Bs} on the entire  {sampled} dataset. Importantly, our framework neither requires nor assumes that $\mathcal{L}(\varphi)$ must be driven to exactly zero. In practice, the synthesis stage only  {minimizes} $\mathcal{L}$ over a finite collection of sampled parameter states under finite computational budget, and it is therefore natural to terminate optimization after a prescribed number of iterations or once $\mathcal{L}(\varphi)$ falls below a chosen tolerance. Consequently, the obtained $\mathcal{B}_{\varphi}$ should be viewed as an empirical  {candidate} certificate whose constraints are enforced only on the sampled data.
However, since NNBC $\mathcal{B}_{\varphi}$ is trained based on a finite set of parameter states, it does not cover the entire set $\Theta $. To overcome this limitation, we establish a probabilistic guarantee that extends the validity of the certificate beyond the training samples with some confidence.

\section{ROBUSTNESS CERTIFICATES VERIFICATION}
\label{sec:prob_guarantee}
The synthesis procedure in Section~\ref{sec:nn_barrier} produces an NNBC $\mathcal{B}_{\varphi}$ and a candidate certified robust radius. However, this data-driven construction enforces the conditions in Definition~\ref{def:barrier_function} only on sampled data and does not by itself guarantee their validity over all states within the proposed radius $\delta_{\mathrm{cert}}$ (or $\delta_{\mathrm{cert}}'$).
To address this generalization gap, we establish a formal probabilistic robustness guarantee for the learned certificate and robust radius beyond the training samples. Specifically, we assume that the learned NNBC $\mathcal{B}_{\varphi}$ is fixed and define
$\Theta_{\mathcal{Z}} \!=\! \bigl\{ \theta \!\in\! \Theta \mid \mathcal{B}_{\varphi}(\theta) \!\leq\! 0 \bigr\}$.
{Given a general admissible perturbation set $\bar{\Lambda}\!\in\!\{\Lambda,\Lambda'\}$, together with its corresponding perturbation matrix $\bar{\Delta}\!\in\!\{\Delta,\Delta'\}$ and certified robust radius $\bar{\delta}_\mathrm{cert}\!\in\!\{\delta_\mathrm{cert},\delta'_\mathrm{cert}\}$, we define a scalar margin $\eta_r$ and functions $q_k:\Theta\!\times\!\bar{\Lambda}\to\mathbb{R}$, for all $k\in[3]$, associated with the conditions in Definition~\ref{def:barrier_function}, as follows:}
\begin{align}
\label{b1}
    &q_1(\theta,\bar{\Delta}) = \big(\mathcal{B}_{\varphi}(\theta)-\eta_r\big)\mathbf{1}_{\Theta _0}, \\
    &q_2(\theta,\bar{\Delta}) = \big(-\mathcal{B}_{\varphi}(\theta)-\eta_r\big)\mathbf{1}_{\Theta _u^{\Delta'}}, \\
        \label{b3} 
        &q_3(\theta,\bar{\Delta})
  = \big(\mathcal{B}_{\varphi}(f{\mbox{\small$(\theta,\Delta)$}}) - \eta_r\big)\,\mathbf{1}_{\Theta_{\mathcal{Z}}}.
\end{align}
{Observe that $q_1$ is independent of the perturbation variable, whereas $q_2$ and $q_3$ depend on $\Delta'$ and $\Delta$, respectively. Nevertheless, for simplicity of notation, we write all three over $\Theta\times\bar{\Lambda}$.}
\subsection{ROBUST CONVEX PROBLEM (RCP)}
To robustly verify the BC conditions \eqref{b1}-\eqref{b3} under all possible poisoning perturbations $\|\bar{\Delta}\|_p \!\leq\! \bar{\delta}_{\mathrm{cert}}$, we formulate an RCP over the only decision variable $\eta_r$, which encodes the worst-case margin for each condition over the entire parameter set~$\Theta$, as follows:
\begin{align}
\label{eq:rcp}
\text{RCP}\!:\!\left\{
\begin{aligned}
\min_{{\eta_r\in\mathbb{R}}} ~ & \eta_r \\
\text{s.t.} ~ 
&q_k(\theta,\bar{\Delta}) \!\le\! 0, {\forall (\theta,\bar{\Delta}) \!\in\! (\Theta\!\times\!\bar{\Lambda})}, \forall k \!\in\! [3].
\end{aligned}
\right.
\end{align}
Since $\mathcal{B}_{\varphi}$ is fixed, the RCP is a robust linear program over the scalar variable $\eta_r$, with the optimal value denoted by $\eta_r^*$. A solution $\eta_r^* \leq 0$ certifies that $\mathcal{B}_{\varphi}$ satisfies the conditions in Definition~\ref{def:barrier_function}, and thus provides an exact robustness certificate for the poisoning attack with the corresponding radius $\bar{\delta}_{\mathrm{cert}}$ with a guarantee 100\%. However, solving this robust linear program is intractable, as the state transition map $f$ is not available under unknown poisoning attacks and the robust problem involves infinitely many constraints due to $\theta$ and $\bar{\Delta}$ belonging to some continuous sets. 
To make this tractable, we relax the RCP into the following chance-constrained problem (CCP): 
\begin{align}
\label{eq:ccp}
\text{CCP}:\left\{
\begin{aligned}
\min_{{\eta_r\in\mathbb{R}}} ~~ & \eta_r \\
\text{s.t.} ~~ 
&\mathbb{P}\left[q_k(\theta,\bar{\Delta}) \!\le\! 0\right]\geq\!1-\epsilon, ~~\forall k \!\in\! [3]
\end{aligned}
\right.
\end{align}
{where the probability is considered across the random outcomes of the joint distribution of $(\theta,\bar{\Delta})$}, and $\epsilon \in (0,1)$ denotes the given violation probability. The goal is to solve the CCP in \eqref{eq:ccp} rather than the RCP in \eqref{eq:rcp}. The CCP optimally discards a constraint subset of probability mass at most $\epsilon$ to maximize objective improvement. However, solving CCP is still challenging since both $\theta$ and $\bar{\Delta}$ lie in continuous spaces. Therefore, we tackle the associated Scenario Convex Problem (SCP).

\subsection{SCENARIO CONVEX PROBLEM (SCP)}
{We approximate the infinitely many constraints using $\hat N$ i.i.d.\ sampled scenarios, which induce the joint sampled sets}
\begin{align*}
&{\mathcal{Z}_1
\!:=\!\{\theta_i\mid \theta_i\!\in\!\Theta_0,\ i\!\in\![\hat N]\},}\\
&{\mathcal{Z}_2
\!:=\!\{(\theta_i,\bar{\Delta}_i)\mid (\theta_i,\bar{\Delta}_i)\!\in\!(\Theta_u^{\Delta'},\bar{\Lambda}),\ i\!\in\![\hat N]\},}\\ 
&{\vartheta'
~\!:=\!\{(\theta_i,\bar{\Delta}_i)\mid (\theta_i, \bar{\Delta}_i)\!\in\!(\Theta,\bar{\Lambda}),\ i\!\in\![\hat N]\}},\\
&{\mathcal{Z}_3= \{ (\theta_i,\bar{\Delta}_i)\mid(\theta_i,\bar{\Delta}_i)\in\vartheta', \mathcal{B}_{\varphi}({\theta}_i)\!\leq\!0\}.}
\end{align*}
{Then, SCP enforces the inequalities only in these sampled scenarios for all $i\!\in\![\hat{N}]$ and $\|\bar{\Delta}_i\|_p \!\le\! \bar{\delta}_\mathrm{cert}$ as follows: }
\begin{align}
\label{eq:scp}
\text{SCP}\!:\!\left\{
\begin{aligned}
\min_{{\eta_s\in\mathbb{R}}} ~ & \eta_s \\
\text{s.t.} ~ 
&q_k(\theta_i,\bar{\Delta}_i) \!\le\! 0, {\forall (\theta_i,\bar{\Delta}_i) \!\in\! \mathcal{Z}_k} , \ \forall k \!\in\! [3].
\end{aligned}
\right.
\end{align}
Let $\eta^*_s$ denote the optimal value of the SCP, resulting in the validated NNBC denoted by $\mathcal{B}_{\varphi}^*$ and its associated certified robust radius, represented by $\bar{\delta}_{\mathrm{cert}}^*$.
Since the SCP replaces the infinite set with finitely many trajectories, it is crucial to assess the generalization of this solution. 
{Hence, we establish a probabilistic guarantee for the scenario-based formulation, which certifies the validity of the constructed BC $\mathcal{B}_{\varphi}$ with some confidence and a violation probability of at most $\epsilon$.}

\subsection{PAC GUARANTEE}
\label{subsec:pac}
To rigorously connect the CCP and SCP, we adopt a probably approximately correct (PAC) guarantee based on Theorem~1 in \cite{calafiore2006scenario}. {Specifically, with a confidence of at least $1\!-\!\beta$, the solution $\eta_s^*$ obtained from the SCP satisfies the chance constraint in the CCP with violation probability of at most $\epsilon$, provided that the number of i.i.d.\ scenarios $\hat{N}$ satisfies}
\begin{align}\label{N_hat}
\hat{N} ~\ge~
\big\lceil \frac{ \ln (\beta) }{\ln(1 - \epsilon) } \big\rceil.
\end{align}
We now present the main theoretical result of this paper. 

\begin{theorem}[Robustness Certificates for Neural Networks]
\label{thm:main_pac_guarantee}
Let $h_{\theta}$ be an ML model, trained on a (potentially) poisoned training dataset $\mathcal{D}_{\mathrm{train}}^\Delta$ and evaluated on a (potentially) poisoned test dataset $\mathcal{D}_{\mathrm{test}}^{\Delta'}$, as described in Section~\ref{sec:prelim}. The model updates adhere to the gradient-based optimization rule $f$ described in \eqref{eq:dtDS}.
Given constants $\alpha \in [0,1]$, $\epsilon \in (0,1)$, and $\beta \in [0,1]$, suppose that an NNBC $\mathcal{B}_{\varphi}^*$ is synthesized from finitely many sampled trajectories as in \eqref{I}-\eqref{U}, yielding a certified train-time robust radius $\delta^*_{\mathrm{cert}}$ (resp. test-time robust radius $\delta'^{*}_{\mathrm{cert}}$).  
Let $\eta_s^* \!\leq\! 0$ be the optimal barrier margin obtained by solving the SCP in \eqref{eq:scp} on $\hat{N}$ i.i.d. samples, with $\hat{N}$ satisfying the bound in \eqref{N_hat}.
Then, with a confidence of at least $1 - \beta$, the learned certificate $\mathcal{B}_{\varphi}^*$ ensures that, for all poisoning perturbations satisfying $\|\Delta\|_p \leq \delta^*_{\mathrm{cert}}$ (resp. $\|\Delta'\|_p \leq \delta'^{*}_{\mathrm{cert}}$), 
the test accuracy degradation remains within the threshold $\alpha$, with the violation probability of at most $\epsilon$.
\end{theorem}

A concise outline of the certification pipeline for poisoning (train-time) attacks is provided in Algorithm~\ref{algorithm}. The procedure consists of two distinct stages: synthesis and verification.
In the synthesis stage, $N$ poisoned roll-outs are generated over the full perturbation budget grid to construct the sampled parameter sets $\mathcal{I}$ (initial states) and $\mathcal{S}/\mathcal{U}$ (safe/unsafe terminal states). Sampling across the full budget range ensures coverage of both regions and enables data-driven learning of the NNBC $\mathcal{B}_\varphi$. Based on these roll-outs, an empirical robustness estimate $\delta_{\mathrm{emp}}$ is first computed and used to initialize $\delta_{\mathrm{cert}}$, after which the NNBC is trained under this initial radius.
In the verification stage, the resulting radius $\delta_{\mathrm{cert}}$ is evaluated using a fresh, disjoint set of $\hat N$ i.i.d. samples again over the full perturbation budget. These samples are used to solve the corresponding SCP and compute $\eta_s^*$, thereby testing whether $\mathcal{B}_\varphi$ separates the safe and unsafe regions under $\delta_{\mathrm{cert}}$. If $\eta_s^* \le 0$, the radius is accepted as $\delta_{\mathrm{cert}}^*$. Otherwise, $\delta_{\mathrm{cert}}$ is reduced and the synthesis--verification procedure is repeated until $\eta_s^* \le 0$ is satisfied.
Note that the certification procedure for test-time attacks mirrors the train-time case, with the only difference that adversarial perturbations are applied at test time rather than during training. See the extended versions for poisoning and evasion attacks in the Appendix~\ref{app:sec_algo}.

\begin{algorithm}[!tb]
\caption{\footnotesize Robustness Certificates against
Poisoning Attacks}
\small
\label{algorithm}
\textbf{Input}: ML Model $h_{\theta}$, gap threshold $\alpha$, number of samples for training NNBC $N$, number of scenarios for solving SCP $\hat{N}$, training horizon $t_\infty$, confidence level $1\!-\!\beta$\\
\textbf{Output}: Trained $\mathcal{B}_\varphi^*$, certified robust radius $\delta_{\mathrm{cert}}^*$, violation probability $\epsilon$.
\vspace{2pt}

\begin{algorithmic}[1]
\STATE Sample $N$ poisoned training trajectories dataset. 
\STATE Train $h_{\theta}$ on each to obtain $\theta_i$ for all $i \in \{1,\ldots, N\}$. Collect terminal parameters ${\theta}_i{\mbox{\small$(t_\infty)$}}$ and label as safe or unsafe using $\mathcal{G}(\mbox{\small$\theta_i(t_\infty)$}) \leq \alpha$; assign ${\theta}_i{\mbox{\small$(0)$}}$ to initial set. \\
\STATE Fix $\delta_{\mathrm{emp}} \gets \max \{ \delta_i \mid \mathcal{G}(\mbox{\small$\theta_i(t_\infty)$}) \leq \alpha \}$ and initialize $\delta_{\mathrm{cert}} \gets \delta_{\mathrm{emp}}$.\\
\STATE Train an NNBC $\mathcal{B}_\varphi$ on collected data to minimize $\mathcal{L}$ and then fix it. 
\STATE Generate $\hat{N}$ fresh i.i.d. samples from the joint distribution of $(\theta, \bar{\Delta})$.\\
\STATE Solve the SCP in equation \eqref{eq:scp} to obtain the margin $\eta_s^*$. 
\STATE If $\eta_s^* > 0$, reduce $\delta_{\mathrm{cert}}$ or increase $N$, and return to Step 2 until $\eta_s^* \leq 0$. Then, compute the minimum violation probability $\epsilon$ from condition~(\ref{N_hat}).
\STATE If there is no $\mathcal{B}_\varphi^*$, $h_\theta$ can not be certified at threshold $\alpha$.\\
\end{algorithmic}
\end{algorithm}

\section{EXPERIMENTAL RESULTS}
\label{sec:experiments}

In this section, we evaluate the effectiveness of our proposed framework and analyze how key design choices impact the certified robustness of the trained model $h_\theta$ under the $\ell_\infty$ and $\ell_2$ train-time threat models. Additional experiments for test-time certification are provided in the Appendix~\ref{app:sec_exp}.

\subsection{ML SETUP} We evaluate our method on four widely used image-classification benchmarks: 
{MNIST}. 
{SVHN}. 
{CIFAR-10}. 
and {CIFAR-100}. 
Robustness is assessed against three representative poisoning strategies during training, Projected Gradient Descent (PGD) (gradient-based input poisoning via projected updates under an $\ell_p$ constraint) \cite{madry2017towards}, Backdoor Attack (BDA) (injecting a fixed trigger pattern so the model learns a targeted misclassification when the trigger is present) \cite{gu2017badnets}, and Bullseye Polytope Attack (BPA) (clean-label feature-collision style poisoning that pulls selected samples toward a target representation) \cite{aghakhani2021bullseye}, and against PGD (iterative white-box test-time perturbations that maximize loss within an $\ell_p$ budget) \cite{madry2017towards} and AutoAttack (AA) (a parameter-free evaluation suite combining complementary attacks such as APGD and FAB to provide a stronger robustness assessment) \cite{croce2020reliable} at test time. The hypothesis class $h_\theta$ spans multiple architectures, including  {MLP},  {CNN}, and  {ResNet}, trained with optimizers  {GD},  {SGD}, and  {Adam}. We provide the experimental configurations and hyperparameters in Table~\ref{tab:configs}, list all model architectures in Table~\ref{tab:model-specs}, and provide further details on the attack models in Table~\ref{tab:attack_models} in the Appendix.

\begin{figure*}[!htb]
    \centering

    \begin{minipage}{0.24\linewidth}
        \centering
        \includegraphics[width=\linewidth]{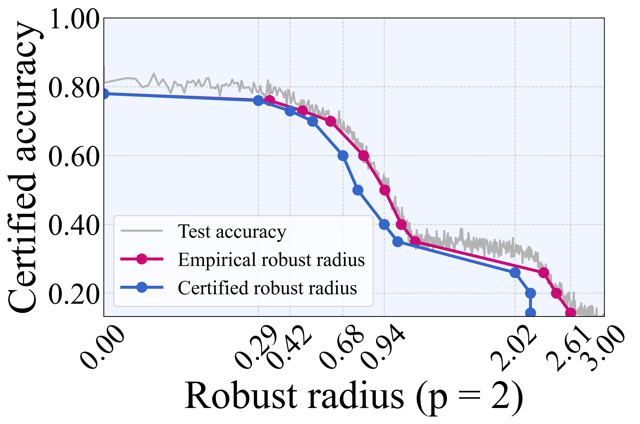}\\[-2pt]
        \small (a) CIFAR10, ResNet, BPA \\($\rho=0.2$)
    \end{minipage}\hfill
    \begin{minipage}{0.24\linewidth}
        \centering
        \includegraphics[width=\linewidth]{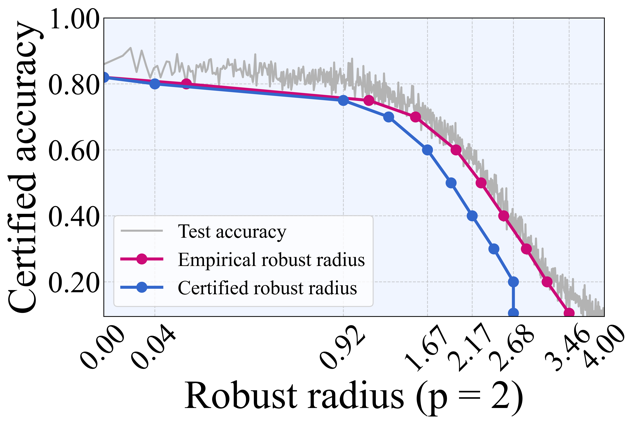}\\[-2pt]
        \small (b) SVHN, ResNet, PGD \\($\rho=0.9$)
    \end{minipage}\hfill
    \begin{minipage}{0.24\linewidth}
        \centering
        \includegraphics[width=\linewidth]{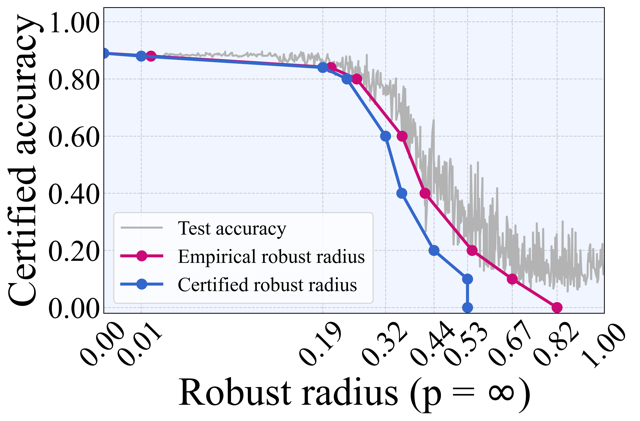}\\[-2pt]
        \small (c) SVHN, MLP, BDA \\($\rho=0.1$)
    \end{minipage}\hfill
    \begin{minipage}{0.24\linewidth}
        \centering
        \includegraphics[width=\linewidth]{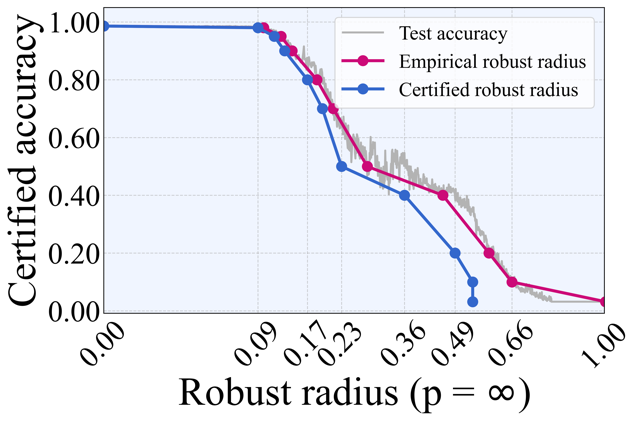}\\[-2pt]
        \small (d) MNIST, CNN, PGD \\($\rho=1$)
    \end{minipage}
\caption{
Certified accuracy ($g_{\mathrm{p}}^*$) versus perturbation magnitude ($\delta$) 
on different settings and poisoning scenarios. Each figure reports the terminal 
test accuracy $g(\theta(t_\infty))$, the empirical robust radius $\delta_\mathrm{emp}$, 
and the certified robust radius $\delta^*_\mathrm{cert}$ obtained using our proposed framework. 
The confidence level is fixed at $1-\beta$, $\beta = 10^{-4}$, across all settings, 
with the corresponding violation probabilities being (a) $0.015$, (b) $0.013$, 
(c) $0.006$, and (d) $0.005$.}
\vspace{-0.2cm}
    \label{fig:certified-accuracy-grid}
\end{figure*}

\begin{table}[t]
\centering
\caption{Certification results for the train-time $\ell_2$ and $\ell_\infty$ attacks and datasets in Figure~\ref{fig:certified-accuracy-grid}. Each row reports the certified accuracy ($g_{\mathrm{p}}^*$), the empirical ($\delta_{\mathrm{emp}}$) and certified ($\delta_{\mathrm{cert}}^*$) radii, BC margin ($\eta_s^*$), and violation probability ($\epsilon$), evaluated at a performance gap threshold $\alpha$ and a confidence level of at least $99.99\%$.}
\renewcommand{\arraystretch}{1.1}
\scalebox{0.65}{
\setlength{\tabcolsep}{4pt} 
\begin{tabular}{c c c c c c c c c c c c c}
\toprule
\textbf{Dataset} & \textbf{Model} & \textbf{Opt} & $\boldsymbol{\mathcal{A}}$ & $\boldsymbol{\rho}$ & {\textbf{$\boldsymbol{p}$}} & \textbf{N} & $\hat{\textbf{N}}$ & $\boldsymbol{g_{\mathrm{p}}^*}$ &
$\boldsymbol{\delta_{\mathrm{emp}}}$ & $\boldsymbol{\delta_{\mathrm{cert}}^*}$ &
$\boldsymbol\eta^*_s$ & $\boldsymbol\epsilon$ \\
\midrule
\multirow{2}{*}{MNIST} & \multirow{2}{*}{CNN}
& \multirow{2}{*}{SGD} & \multirow{2}{*}{PGD} & \multirow{2}{*}{1} & \multirow{2}{*}{$\infty$}& \multirow{2}{*}{4000} & \multirow{2}{*}{1800}
& 0.90 & 0.14 & 0.13 & -0.01 & \multirow{2}{*}{0.005} \\
&  && & & & & & 0.80 & 0.18 & 0.16 & -0.01 & \\
\midrule
\multirow{4}{*}{SVHN} & \multirow{2}{*}{ResNet}
& \multirow{2}{*}{Adam} & \multirow{2}{*}{PGD} & \multirow{2}{*}{0.9}  & \multirow{2}{*}{2}& \multirow{2}{*}{3000} & \multirow{2}{*}{700}
& 0.75 & 1.21 & 0.92 & -0.11 & \multirow{2}{*}{0.013} \\
&  &&& &  & & & 0.60 & 2.0 & 1.67 & -0.05 & \\
\cmidrule{2-13}
& \multirow{2}{*}{MLP}& \multirow{2}{*}{SGD} & \multirow{2}{*}{BDA} & \multirow{2}{*}{0.1}  & \multirow{2}{*}{$\infty$}& \multirow{2}{*}{4000} & \multirow{2}{*}{1500}
& 0.80 & 0.25 & 0.23 & -0.02 & \multirow{2}{*}{0.006} \\
& && & &  & & & 0.60 & 0.35 & 0.32 & -0.03 & \\\midrule
\multirow{2}{*}{CIFAR-10}
& \multirow{2}{*}{ResNet}& \multirow{2}{*}{Adam} & \multirow{2}{*}{BPA} & \multirow{2}{*}{0.2} & \multirow{2}{*}{$\infty$} & \multirow{2}{*}{2000} & \multirow{2}{*}{600}
& 0.70 & 0.61 & 0.52 & -0.01 & \multirow{2}{*}{0.015} \\
& && & &  & & & 0.60 & 0.84 & 0.68 & -0.02 & \\
\midrule
\multirow{4}{*}{CIFAR-100} & \multirow{4}{*}{ResNet} & \multirow{4}{*}{Adam} & \multirow{2}{*}{PGD} & \multirow{2}{*}{0.6} & \multirow{2}{*}{2} & \multirow{2}{*}{800} & \multirow{2}{*}{200}
& 0.70 & 1.07 & 0.95 & 0.00 & \multirow{2}{*}{0.045} \\
 &  &  &  &  &  &  & 
& 0.60 & 1.75 & 1.62 & 0.00 &  \\
\cmidrule{4-13}
 &  &  & \multirow{2}{*}{BPA} & \multirow{2}{*}{0.2} & \multirow{2}{*}{$\infty$} & \multirow{2}{*}{1000} & \multirow{2}{*}{300}
& 0.70 & 0.52 & 0.43 & -0.01 & \multirow{2}{*}{0.030} \\
 &  &  &  &  &  &  &
& 0.60 & 0.78 & 0.72 & -0.01 &  \\
\bottomrule
\end{tabular}
}
\label{tab:certification}
\end{table}

\subsection{CERTIFICATION RESULTS}
We present representative results of different combinations of $\ell_2$ and $\ell_\infty$  {train-time} attacks under poisoning ratio $\rho$ ranging from $0.1$ to $1$ and datasets, with a confidence level of $99.99\%$ in Table~\ref{tab:certification} and Figure~\ref{fig:certified-accuracy-grid} (see Table~\ref{tab:configs} and Figure~\ref{fig:All_certified-accuracy-grid} for all the results).    
We denote the  {certified accuracy} by $g_{\mathrm{p}}^*$ which is computed by $g_{\mathrm{p}}^*\!=\!g_{\mathrm{c}}\!-\!\alpha$ as in Definition~\ref{def:deg}. Note that the robustness specification is flexible: fixing the degradation threshold $\alpha$ is equivalent to prescribing a target certified test accuracy, since \eqref{eq:deg} is linear in $g_{\mathrm{p}}$. 
 {Non-trivial certificates} are obtained in all settings. 
Exemplary, for SVHN, at $g_{\mathrm{p}}^* \!=\! 0.75$ under PGD ($\ell_2$), our framework certifies robust radii up to $\delta^{*}_{\mathrm{cert}} \!=\! 0.92$ even when the poisoning ratio is as high as $\rho\!=\!0.9$.
All SCP margins $\eta_s^*$ are non-positive, certifying satisfaction of the sampled barrier constraints, and the violation probability $\epsilon$ remains below $0.02$ in most configurations (Table~\ref{tab:certification}).
Importantly, our results yield tight certificates, with the certified robust radius $\delta_{\mathrm{cert}}$ consistently close to the empirical robust radius $\delta_{\mathrm{emp}}$. While $\delta_{\mathrm{cert}} \!\le\! \delta_{\mathrm{emp}}$ holds by construction, tightness is largely dictated by how regularly the model degrades under poisoning: when the test accuracy $g$ decreases smoothly rather than oscillates, the NNBC generalizes better and the certified radii closely track their empirical counterparts. See Appendix~\ref{app:discussion} for a detailed analysis of the experimental results.

\begin{table}[!tb]
\centering
\caption{Comparison of certified robust radii ($\delta_{\mathrm{cert}}^*$) obtained by our proposed framework versus the RAB baseline \cite{RAB} under Backdoor Attacks (BDA), where $\textbf{RT}$ represents the simulation runtime (in minutes). Results are reported at different target certified accuracy levels ($g_{\mathrm{p}}^*$). ``NA'' indicates that RAB failed to certify any valid radius. Our method consistently yields larger certified radii that are tighter to the empirical upper bound ($\delta_{\mathrm{emp}}$).}
\label{tab:framevsRAB}
\renewcommand{\arraystretch}{1.1}
\scalebox{0.65}{
\setlength{\tabcolsep}{6pt}
\begin{tabular}{
c c c c c c c c c c c
}
\toprule
\multirow{2}{*}{\textbf{Dataset}} & \multirow{2}{*}{\textbf{Optimizer}} & \multirow{2}{*}{$\boldsymbol{\mathcal{A}}$} & \multirow{2}{*}{$\boldsymbol{\rho}$} & \multirow{2}{*}{{$\boldsymbol{p}$}} &
\multirow{2}{*}{$\boldsymbol{g_{\mathrm{p}}^*}$} & \multirow{2}{*}{$\boldsymbol{\delta_{\mathrm{emp}}}$} &
\multicolumn{2}{c}{\textbf{RAB}} &
\multicolumn{2}{c}{\textbf{Ours}} \\
\cmidrule(lr){8-9}\cmidrule(lr){10-11}
& & & & & & &
$\boldsymbol{\delta_{\mathrm{cert}}^*}$ & \textbf{RT} &
$\boldsymbol{\delta_{\mathrm{cert}}^*}$ & \textbf{RT} \\
\midrule

\multirow{3}{*}{MNIST}
& \multirow{3}{*}{SGD} & \multirow{3}{*}{BDA} & \multirow{3}{*}{0.15} & \multirow{3}{*}{$\infty$}
& 0.90 & 0.08 & NA   & -- & 0.08 & 6 \\
&      &      &      &      & 0.80 & 0.16 & 0.10 & 57 & 0.15 & 9 \\
&      &      &      &      & 0.60 & 0.22 & 0.14 & 51 & 0.19 & 17 \\
\midrule

\multirow{3}{*}{SVHN}
& \multirow{3}{*}{SGD} & \multirow{3}{*}{BDA} & \multirow{3}{*}{0.1} & \multirow{3}{*}{2}
& 0.80 & 0.09 & NA   & -- & 0.06 & 15 \\
&      &      &      &      & 0.60 & 0.12 & 0.07 & 62 & 0.11 & 22 \\
&      &      &      &      & 0.40 & 0.16 & 0.08 & 69 & 0.14 & 12 \\
\midrule

\multirow{3}{*}{CIFAR-10}
& \multirow{3}{*}{Adam} & \multirow{3}{*}{BDA} & \multirow{3}{*}{0.1} & \multirow{3}{*}{$\infty$}
& 0.50 & 0.05 & NA   & -- & 0.04 &48 \\
&      &      &      &      & 0.40 & 0.09 & NA   & -- & 0.07 & 66 \\
&      &      &      &      & 0.30 & 0.15 & 0.05 & 174 & 0.12 & 51 \\
\bottomrule
\end{tabular}
}
\end{table}

\begin{figure}[!tb] 
    \centering  \includegraphics[width=0.7\linewidth]{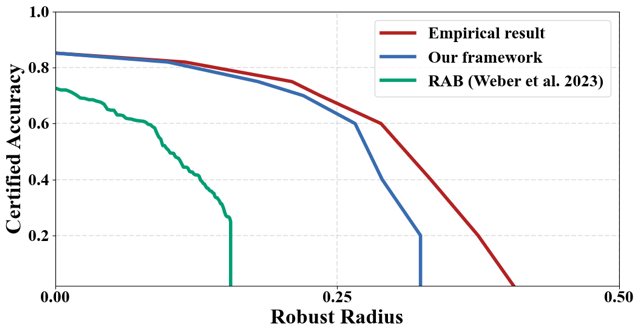}
    \caption{Comparing the result of our framework and RAB on SVHN under test-time BDA with $\ell_\infty$ attack. Our results consistently yield higher certified robustness than RAB.}
    \label{fig:ours-vs-rab}
\end{figure}

\subsection{COMPARISON WITH RELATED WORKS}
We compare our results with RAB, a randomized smoothing–based certified defense against evasion and backdoor attacks \cite{RAB}. As shown in Figure~\ref{fig:ours-vs-rab},  {we consistently achieve stronger and tighter guarantees than RAB}. Notably, our results more closely match the empirical robustness (see Table~\ref{tab:framevsRAB} for more results). 
Other feature-poisoning certificates are less suitable for direct comparison: BagFlip \cite{zhang2022bagflip} is restricted to $\ell_0$ corruptions; ensemble-based methods permit unbounded perturbations \cite{levine2021deep, wang2022improved}; and model-specific approaches tailored to neural networks either apply only to infinite-width graph neural networks \cite{gosch2025provable} or impose unrealistic restrictions on training and model choice \cite{sosni2024abstractgradient}.
More experiments, including diverse train- and test-time attack scenarios and a runtime analysis, are provided in Appendix~\ref{app:sec_exp}.

\subsection{EFFECT OF SAMPLING DENSITY}
\begin{figure}[!h]
    \centering
    \begin{minipage}{0.45\linewidth}
        \centering
        \includegraphics[width=0.9\linewidth]{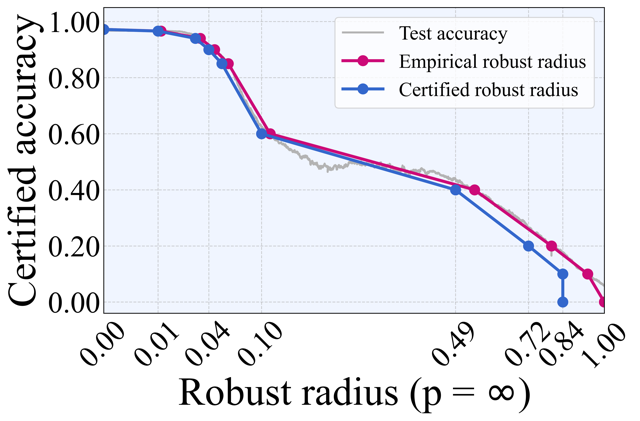} \\
       \tiny (a) Sparse $\alpha$-sampling (10 thresholds)
    \end{minipage}
    \hfill
    \begin{minipage}{0.49\linewidth}
        \centering
        \includegraphics[width=0.9\linewidth]{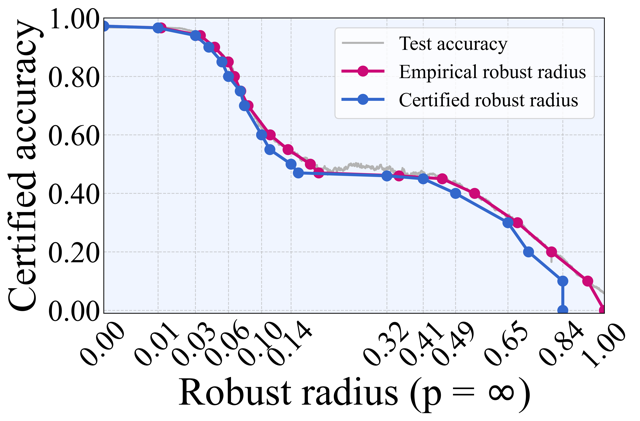} \\
        \tiny (b) Dense $\alpha$-sampling (20 thresholds)
    \end{minipage}
    \caption{Effect of $\alpha$-sampling density on empirical and certified robust-radius curves for MNIST, MLP, PGD, train-time poisoning. }
    \label{fig:accurate}
\end{figure}

In some configurations, the empirical test-accuracy curve appears to fall below the certified robust-radius curves, which is theoretically inconsistent. This artifact is caused by an insufficient sampling of $\alpha$ thresholds when computing $\delta_{\mathrm{cert}}$, leading to interpolation errors. The fidelity of both empirical and certified robust-radius curves therefore depends on the density of these evaluation points. Figure~\ref{fig:accurate} illustrates this effect by increasing the number of sampled $\alpha$ values from $10$ in (a) to $20$ in (b). The curves in (b) are smoother and more faithful to the underlying robustness profile, and no longer exhibit anomalous crossings.

\vspace{-0.5cm}
\subsection{COMPUTATIONAL COST AND SCALABILITY}
Computational cost is still an open challenge in robustness certification: formal guarantees typically require either restrictive assumptions or extensive computation over rich adversarial/scenario spaces. Our framework follows this general trade-off.
Its main cost comes from generating perturbed training roll-outs for NNBC synthesis and an additional disjoint scenario set for verification. Here, $\hat N$ is fixed by the target PAC parameters via~\eqref{N_hat}, while $N$ governs the trade-off between coverage, certificate tightness, and runtime. Empirically, $N \!\geq\! 1000$ is usually sufficient for reliable NNBC synthesis and SCP validation, while larger $N$ can further reduce conservatism at higher computational cost.
End-to-end runtimes are reported in Table~\ref{tab:configs}, with a direct comparison to RAB in Table~\ref{tab:framevsRAB}. Unlike our model-level certification framework, RS/RAB certifies each test sample separately and then aggregates the resulting point-wise certificates over the full test set. Its total runtime therefore scales approximately with the number of test samples (up to parallelization), which helps explain why, under matched settings, it can exceed ours.

\vspace{-0.2cm}
\section{CONCLUSION}
This paper proposed a formal robustness certification framework for neural networks under data poisoning and evasion attacks. We modeled gradient-based training as a discrete-time dynamical system and cast robustness as a safety verification problem in parameter space, using barrier certificates to certify an $\ell_p$-bounded robust radius under which test-accuracy degradation remains below a prescribed threshold. To handle high-dimensional and unknown training dynamics, we parameterized the barrier certificate as a neural network learned from poisoned trajectories and verified it via a scenario convex program, yielding probabilistic guarantees on the certified radius. The framework is unified, model-agnostic, and applicable to both train-time and test-time feature perturbations. Experiments on MNIST, SVHN, and CIFAR show non-trivial certified budgets and certified radii close to empirical robustness. Future work includes extensions beyond $\ell_p$ threat models, lower certification cost, and time-varying training hyperparameters.

\section*{APPENDIX} \label{Appendix}

\renewcommand{\thesection}{\Alph{section}}
For more clarity, key symbols and quantities used throughout the paper are summarized in Table~\ref{tab:frame-symbols}.

\begin{table}[!t]
\caption{Summary of key symbols and definitions.}
\label{tab:frame-symbols}
\centering
\renewcommand{\arraystretch}{1.453}
\scalebox{0.83}{
\setlength{\tabcolsep}{3pt}
\begin{tabular}{ll}
\toprule
\textbf{Symbol} & \textbf{Definition / Scope} \\
\midrule
\multicolumn{2}{l}{\textbf{Threat Model \& Perturbation}} \\
$p$ & Norm type of the admissible perturbation set \\&($p=2$ or $p=\infty$ in the experiments) \\
$\delta$, $\delta'$ & Max perturbation magnitude for train/test time \\
$\rho$, $\rho'$ & Fraction of corrupted samples in train/test time \\
$\Delta$, $\Delta'$ & Feature-space perturbation matrices for train/test data \\
$\bar{\Delta}$ & Generic perturbation matrix, either $\Delta$ or $\Delta'$ \\
$\Lambda$, $\Lambda'$ & Admissible perturbation sets for train/test time \\
$\bar{\Lambda}$ & Generic admissible perturbation set, either $\Lambda$ or $\Lambda'$ \\
$\bar{\delta}_{\mathrm{cert}}$ & Generic certified radius, either $\delta_{\mathrm{cert}}$ or $\delta'_{\mathrm{cert}}$ \\
$\mathcal{D}_{\mathrm{train}}^\Delta$ & Poisoned training dataset \\
$\mathcal{D}_{\mathrm{test}}^{\Delta'}$ & Perturbed test dataset \\
$\mathbb{P}$ & Probability measure over sampled scenarios $(\theta,\bar{\Delta})$ \\

\midrule
\multicolumn{2}{l}{\textbf{Training Dynamics \& Parameters}} \\
$\Theta$ & Parameter space \\
$\Theta_0$ & Set of admissible initial model parameters \\
$\theta(0)$, $\theta(t)$ & Model parameters at initialization and iteration $t$ \\
$t_\infty$ & Finite terminal training horizon used for certification \\
$f$ & Gradient-based update rule as in~\eqref{eq:dtDS} \\
$\mathfrak{S}$ & Discrete-time dynamical system modeling training \\

\midrule
\multicolumn{2}{l}{\textbf{Safety Criterion and Robust Radii}} \\
$g(\theta;\mathcal{D}_{\mathrm{test}}^{\Delta'})$ & Test accuracy of $h_\theta$ on a possibly perturbed test set \\
$g_{\mathrm{c}}$, $g_{\mathrm{p}}$ & Clean/reference and attacked test accuracies used in $\mathcal{G}$ \\
$\mathcal{G}(\mbox{\small$\theta$})$ & Test degradation gap: $\mathcal{G}=g_{\mathrm{c}}-g_{\mathrm{p}}$ \\
$\alpha$ & Test-accuracy degradation threshold \\
$\Theta_s^{\Delta'}$, $\Theta_u^{\Delta'}$ & Safe/unsafe parameter sets induced by $\mathcal{G}(\theta)\leq\alpha$ / $>\alpha$ \\
$\delta_{\mathrm{emp}}, \delta_{\mathrm{emp}}'$ & Empirical train/test-time robust radii \\
$\delta_{\mathrm{cert}}^*$, $\delta_{\mathrm{cert}}'^*$ & Certified train/test-time robust radii after verification \\
$g_{\mathrm{p}}^*$ & Certified accuracy, $g_{\mathrm{p}}^*=g_{\mathrm{c}}-\alpha$ \\

\midrule
\multicolumn{2}{l}{\textbf{Neural Barrier Certificate (NNBC)}} \\
$\mathcal{B}$ & Barrier certificate \\
$\mathcal{B}_\varphi(\theta)$ & Neural network-based barrier certificate parameterized by $\varphi$ \\
$\mathcal{B}_\varphi^*$ & Verified NNBC associated with the certified radius \\
$\mathcal{I}$ & Sampled initial parameter set \\
$\mathcal{S}$, $\mathcal{U}$ & Sampled safe and unsafe terminal parameter sets \\
$\vartheta$ & Collected NNBC training set $\mathcal{I}\cup\mathcal{S}\cup\mathcal{U}$ \\
$\mathcal{Z}$ & Sampled barrier sublevel set used in the loss \\
$\Theta_{\mathcal{Z}}$ & Barrier sublevel set $\{\theta\in\Theta:\mathcal{B}_{\varphi}(\theta)\leq0\}$ \\
$\mathcal{L}(\varphi)$ & Loss function for NNBC synthesis in~\eqref{Loss_NNBC} \\
$N$ & Number of sampled trajectories used for NNBC synthesis \\

\midrule
\multicolumn{2}{l}{\textbf{Scenario Certification (RCP / CCP / SCP)}} \\
$q_1,q_2,q_3$ & Constraint functions encoding the BC conditions \\
$\eta_r$ & Margin variable in the robust convex problem (RCP) \\
$\eta_s$ & Margin variable in the scenario convex problem (SCP) \\
$\eta_s^*$ & Optimal SCP margin used for certification \\
$\mathcal{Z}_1,\mathcal{Z}_2,\mathcal{Z}_3$ & Scenario sets for initial, unsafe, and forward BC conditions \\
$\hat{N}$ & Number of i.i.d. scenarios used for SCP verification \\
$\epsilon$ & Allowed violation probability over unseen scenarios \\
$\beta$ & PAC confidence parameter, yielding confidence $1-\beta$ \\
\bottomrule
\end{tabular}
}
\end{table}

\subsection{PROOF OF THEOREM~\ref{thm:safety-guarantee}} \label{app:thm_pf_safety-guarantee} 
\begin{proof} By Definitions~\ref{def:safety_specific} and~\ref{def:barrier_function}, the zero-level set of the barrier, $\{\theta\in\mathbb{R}^d \mid \mathcal{B}(\theta)=0\}$, separates the safe region $\Theta_s^{\Delta'} := \{\theta\in\mathbb{R}^d \mid\mathcal{G}(\mbox{\small$\theta$}) \le \alpha\}$ from the unsafe region $\Theta_u^{\Delta'} := \{\theta\in\mathbb{R}^d \mid\mathcal{G}(\mbox{\small$\theta$})  > \alpha\}$.\\
\noindent
\textbf{(1)} 
Without loss of generality, because the parameters at \(t=0\) are randomly initialized with small magnitudes—hence untrained and uninfluenced by data—the model’s predictions are essentially random. Its test accuracy is therefore at chance level, whether evaluated on clean or poisoned inputs. Consequently, the initialization gap satisfies \(\mathcal{G}(\theta(0)) \approx 0\), which is negligible relative to any admissible threshold \(\alpha\). By Definition~\ref{def:safety_specific}, it follows that \(\theta(0)\in\Theta_s^{\Delta'}\).
 \\
\noindent
\textbf{(2)} 
By condition~\eqref{B0}, the initial model parameters $\theta{\mbox{\small$(0)$}} \in \Theta _0$ always satisfy $\mathcal{B}(\theta{\mbox{\small$(0)$}}) \leq 0$. This aligns with (1). So training begins inside (or on the boundary of) the barrier zero sublevel set.\\
\noindent
\textbf{(3)} Suppose that at iteration $t$, $\mathcal{B}(\theta(t)) \leq 0$. If $\theta(t) \in \Theta _u^{\Delta'}$, then condition~\eqref{Bu} implies $\mathcal{B}(\theta(t)) > 0$, which is a contradiction. Therefore, $\theta(t)$ must lie in $\Theta _s^{\Delta'}$.\\
\noindent
\textbf{(4)} By condition~\eqref{Bs}, for any admissible poisoning with $\|\Delta\|_p \leq \delta_ \mathrm{cert}$, if $\mathcal{B}(\theta(t)) \le 0$, then the next state $\theta(t+1)$ also satisfies $\mathcal{B}(\theta(t+1)) \le 0$ and the zero sub-level set $\{\theta\in\mathbb{R}^d\,\,|\,\,\mathcal{B}(\theta)\le 0\}$ is forward invariant for the training dynamics under all admissible $\Delta$.\\
\noindent
\textbf{(5)} From (2)–(4) we conclude that once the training starts inside the safe region, the barrier condition guarantees that $\mathcal{B}(\theta(t)) \le 0$ holds for every iteration $t$. In particular, at the terminal time $t = t_\infty$ we have $\mathcal{B}(\theta(t_\infty)) \le 0$, which by the separation property in (3) implies that $\theta(t_\infty) \in \Theta_s^{\Delta'}$. By Definition~\ref{def:deg}, this means that the accuracy gap at convergence satisfies $\mathcal{G}(\mbox{\small$\theta(t_\infty)$})  \le \alpha$, that is, the trajectory remains in $\Theta _s^{\Delta'}$ for all $t$ and never enters $\Theta _u^{\Delta'}$. For test-time perturbations \(\Delta'\), observe that they do not alter the training dynamics and only affect the accuracy at evaluation. Hence, the same separation argument applies: the terminal parameters remain in \(\Theta_s^{\Delta'}\) for all \(\Delta'\) with \(\|\Delta'\|_p \le \delta'_{\mathrm{cert}}\).
\\
\noindent
\textbf{(6)} Hence, $\delta_{\mathrm{cert}}$ (resp. $\delta'_{\mathrm{cert}}$) serves as a certified train-time (resp. test-time) robust radius, guaranteeing that under worst-case admissible perturbations the degradation in test accuracy at convergence is bounded by~$\alpha$.
\end{proof}

\subsection{PROOF OF THEOREM~\ref{thm:main_pac_guarantee}}
\begin{proof}
Let \(\mathbb{P}\) denote a probability measure on the product space \(\Theta \times \bar{\Lambda}\), where \(\Theta\) is the parameter space of the ML model and \(\bar{\Lambda}\) denotes the general set of admissible poisoning perturbation matrices \(\bar{\Delta}\). Our goal is to certify, with high confidence, that the trained model \(h_\theta\) satisfies safety and accuracy constraints even under worst-case poisoning. In the main text, this objective is posed as a robust constrained program (RCP) in \eqref{eq:rcp}, where the constraints must hold for all admissible perturbations.
Since solving the RCP is generally intractable—owing to its dependence on the full uncertainty space and the absence of a closed form for \(f\)—we relax it to a chance-constrained problem (CCP). The CCP permits violations on at most an \(\epsilon\) fraction of the uncertainty set, which is acceptable from a probabilistic safety perspective. In this formulation, we define the violation probability as
\begin{align}
\mathbb{V}(\eta) \!:=\! \mathbb{P} \left[(\theta,\bar{\Delta})\!\in\! \Theta\!\times\! \bar{\Lambda}\!
: \exists k \!\in\![3] \text{ st. } q_k(\theta, \bar{\Delta}) \!> \!\eta \right].
\end{align}
This quantity is the central object of the CCP, capturing the probability that the BC margin \(\eta\) is violated under a random poisoning scenario. We say \(\eta\) is \(\epsilon\)-feasible if \(\mathbb{V}(\eta)\le \epsilon\), i.e., the CCP holds with probability at least \(1-\epsilon\).

To solve the CCP in practice, we approximate it by the SCP in \eqref{eq:scp}, which replaces the probabilistic constraint with empirical constraints over \(\hat N\) i.i.d. scenarios drawn from \(\mathbb{P}\). Concretely, the SCP seeks a margin \(\eta\) such that \eqref{eq:scp} is satisfied.
Let \(\eta_s^*\) be the SCP solution constructed from the sample set
\(\{(\theta_i,\bar{\Delta}_i)\}_{i=1}^{\hat N}\sim \mathbb{P}\).
By the scenario framework of \cite{calafiore2006scenario}, under standard assumptions (e.g., uniqueness and measurability of the solution), the probability that \(\eta_s^*\) violates the original CCP constraint is bounded as
\begin{align}
\mathbb{P}^{\hat N}\!\left(\mathbb{V}(\eta_s^*)>\epsilon\right)
\;\le\;
\sum_{k=0}^{R-1} \binom{\hat N}{k}\,\epsilon^{k}\,(1-\epsilon)^{\hat N-k},
\end{align}
where \(\mathbb{P}^{\hat N}=\mathbb{P}\times\cdots\times\mathbb{P}\) (taken \(\hat N\) times) is the product measure on the full multi-sample \(\bar{\Lambda}\), and \(R\) denotes the number of support constraints of the SCP.

In our setting, the trained NNBC \(\mathcal{B}_\varphi\) is fixed in \eqref{eq:scp}, and the SCP has a single decision variable (the scalar margin \(\eta_s\)). Hence the maximal number of support constraints is \(R=1\). Substituting \(R=1\) into the scenario bound yields
\begin{align}\label{N_hat_app}
\mathbb{P}^{\hat N}\!\left(\mathbb{V}(\eta_s^*)>\epsilon\right)\le (1-\epsilon)^{\hat N}.
\end{align}
To make this failure probability at most \(\beta\), it suffices to require
\((1\!-\!\epsilon)^{\hat N}\!\le\! \beta\), i.e.
$\hat N \!\ge\! \frac{\ln \beta}{\ln(1-\epsilon)}$,
equivalently,
$\hat N \!\ge\! \big\lceil \tfrac{\ln \beta}{\ln(1-\epsilon)} \big\rceil \text{ for integer } \hat N$.
Therefore, if the number of sampled scenarios \(\hat N\) satisfies this condition, then with probability at least \(1\!-\!\beta\) (over the draw of \(\bar{\Lambda}\)), the SCP solution \(\eta_s^*\) is \(\epsilon\)-feasible for the CCP and thus approximates the RCP by certifying safety and test-accuracy constraints on all but an \(\epsilon\)-fraction of poisoning scenarios drawn from \(\mathbb{P}\).
Finally, if \(\eta_s^*<0\) for some poisoning radius \(\delta^*_{\mathrm{cert}}\) (or test-time radius \(\delta'^*_{\mathrm{cert}}\)), we conclude that, with a confidence at least \(1\!-\!\beta\), for all poisoning perturbation matrices \(\Delta\) satisfying \(\|\Delta\|_p \le \delta^*_{\mathrm{cert}}\) (resp. for test-time \(\Delta'\) with \(\|\Delta'\|_p \le \delta'^*_{\mathrm{cert}}\)), the terminal parameters remain in the certified safe set and \(\mathcal{G} \le \alpha\), except on an \(\epsilon\)-fraction of cases.
This shows how the proposed framework inherits these guarantees through this layered connection, completing the proof.
\end{proof}
\vspace{-0.1cm}
\subsection{ALGORITHMS} \label{app:sec_algo}
\begin{figure}[!b] 
\centering\includegraphics[width=0.99\linewidth]{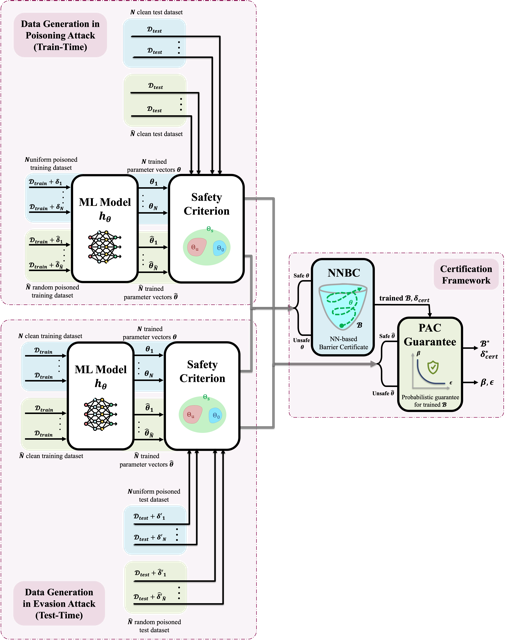}
\vspace{-0.5cm}
\caption{Overview of the proposed framework. The left panel shows data generation under train-time poisoning or test-time evasion attacks. For each perturbation level, the model $h_\theta$ is trained on perturbed datasets to generate two disjoint sets of parameter vectors, $\theta$ and $\hat{\theta}$. A safety criterion is then applied to label each parameter vector as safe or unsafe. The set $\theta$ is used to train the NNBC $\mathcal{B}_\varphi$, while $\hat{\theta}$ is used to verify $\mathcal{B}_\varphi$ through scenario-based PAC analysis. The right panel shows the certification stage, which returns a certified NNBC $\mathcal{B}_{\phi}^*$, the corresponding robust radius $\delta^*_{\mathrm{cert}}$ or $\delta_{\mathrm{cert}}^{\prime *}$, and a probabilistic guarantee with violation probability at most $\epsilon$ and confidence at least $1-\beta$.}
    \label{fig:frame_both}
\end{figure}

We summarize the certification procedure in Figure~\ref{fig:frame_both}, which illustrates the overall workflow. In addition, Algorithms~\ref{alg1} and ~\ref{alg2} describe the certification process under train-time and test-time poisoning settings, respectively. These procedures detail how the NNBC is trained and verified using disjoint parameter sets to provide valid robustness guarantees.

\vspace{-0.1cm}

\begin{algorithm}
\caption{\footnotesize Robustness Certificates for an ML Model $h_{\theta}$ against Data Poisoning Attacks}
\small
\vspace{5pt}
\textbf{Input} Clean train and test datasets $\mathcal{D}_{\text{train}}$ and $\mathcal{D}_{\text{test}}$, model $h_{\theta}$, gap threshold $\alpha$, training horizon $t_\infty$, norm $p$, number of samples for NNBC $N$, number of samples for SCP $\hat{N}$, step size $d_\delta$, confidence level $1-\beta\in[0,1]$, poison ratio $\rho\in[0,1]$, max iterations $T$;\\
\textbf{Output} Certified robust radius~$\delta_{\mathrm{cert}}^*$, NNBC~$\mathcal{B}_\varphi^*$, violation probability~$\epsilon$;\\
\hrule
\vspace{0.5em}
\textbf{Step 1 - Data Generation }
\vspace{0.5em}
\hrule
\vspace{0.5em}
\begin{algorithmic}[1]
\FOR{$i = 1$ to $N$}
    \STATE Initialize ${\theta}_i{\mbox{\small$(0)$}} \in \Theta _0$ and add to $\mathcal{I}$;
    \STATE Sample poisoning level $\delta_i$;
    \STATE Create $\rho$-ratio poisoned dataset $\mathcal{D}_{\text{train}}^{\Delta_i}$ with $\|\Delta_i\|_p\!=\! \delta_i$;
    \FOR{$j = 1$ to $t_\infty$}
        \STATE Train $h_{\theta}$ on $\mathcal{D}_{\text{train}}^{\Delta_i}$ to obtain ${\theta}_i(j)$;
    \ENDFOR
    \STATE Evaluate test accuracy $\mathcal{G}(\mbox{\small$\theta_i(t_\infty)$})$ on $\mathcal{D}_{\text{test}}$ as in definition~\ref{def:deg};
    \IF{$\mathcal{G}(\mbox{\small$\theta_i(t_\infty)$}) \leq \alpha$}
        \STATE Add ${\theta}_i{\mbox{\small$(t_\infty)$}}$ to $\mathcal{S}$;
    \ELSE
        \STATE Add ${\theta}_i{\mbox{\small$(t_\infty)$}}$ to $\mathcal{U}$;
    \ENDIF
\ENDFOR
\STATE Set $\vartheta\gets \mathcal{I} \cup \mathcal{U} \cup \mathcal{S}$;
\STATE Compute $\delta_{\mathrm{emp}} \gets \max \left\{ \delta_i \mid \mathcal{G}({\theta}_i{\mbox{\small$(t_\infty)$}}) \leq \alpha \right\}$;
\RETURN $\mathcal{I}, \mathcal{U}, \vartheta, \delta_{\mathrm{emp}}$;
\end{algorithmic}

\vspace{2em}
\hrule
\vspace{0.5em}
\textbf{Step 2 - Certification Framework}
\vspace{0.5em}
\hrule
\vspace{0.5em}

\begin{algorithmic}[1]
\STATE Generate $N$ trajectories to form $\mathcal{I}, \mathcal{U}, \vartheta$ along with corresponding empirical robust radius $\delta_{\mathrm{emp}}$ and generate $\hat{N}$ i.i.d. samples to form $\mathcal{Z}_1, \mathcal{Z}_2, \vartheta'$;
\STATE Initialize $\delta_{\mathrm{cert}} \gets \delta_{\mathrm{emp}}$, NNBC $\mathcal{B}_\varphi$, and counter $k \gets 0$;
\WHILE{$\delta_{\mathrm{cert}} > 0$}
    \WHILE{$\mathcal{L} \neq 0$ \textbf{and} $k < T$}
        \STATE Train $\mathcal{B}_\varphi$ using $\mathcal{I}, \mathcal{U}, \vartheta$ with loss $\mathcal{L}$ as in \eqref{Loss_NNBC};
        \STATE Update $\mathcal{L}$ and increment $k \gets k + 1$;
        \IF{$\mathcal{L} \neq 0$ \textbf{and} $k\geq T$}
            \STATE Decrease radius: $\delta_{\mathrm{cert}} \gets \delta_{\mathrm{cert}} - d_\delta$ \textbf{break};
        \ELSIF{$\mathcal{L} = 0$}
            \STATE Solve SCP as in \eqref{eq:scp} using $\mathcal{Z}_1, \mathcal{Z}_2, \vartheta'$ to obtain margin $\eta_s^*$;
            \IF{$\eta_s^* > 0$}
                \STATE Decrease radius: $\delta_{\mathrm{cert}} \gets \delta_{\mathrm{cert}} - d_\delta$;
                \STATE (Optional: Increase $N$ );
            \ELSE
                \STATE $\delta_{\mathrm{cert}}^* \gets \delta_{\mathrm{cert}}$;
                \STATE $\mathcal{B}_\varphi^* \gets \mathcal{B}_\varphi$;
                \STATE Compute $\epsilon$ from $\hat{N} = \left\lceil \frac{\ln(\beta)}{\ln(1 - \epsilon)} \right\rceil$;
                \STATE \textbf{break}
            \ENDIF
        \ENDIF
    \ENDWHILE
\ENDWHILE
\RETURN $\mathcal{B}_\varphi^*, \delta_{\mathrm{cert}}^*, \epsilon$
\end{algorithmic}
\label{alg1}
\end{algorithm}

\begin{algorithm}
\caption{\footnotesize Robustness Certificates for an ML Model $h_{\theta}$ against Evasion Attacks}
\small
\vspace{1em}
\textbf{Input} Clean train and test datasets $\mathcal{D}_{\text{train}}$ and $\mathcal{D}_{\text{test}}$, model $h_{\theta}$, gap threshold $\alpha$, training horizon $t_\infty$, norm $p$, number of samples for NNBC $N$, number of samples for SCP $\hat{N}$, step size $d_\delta$, confidence level $1-\beta\in[0,1]$, poison ratio $\rho'\in[0,1]$, max iterations $T$;\\
\textbf{Output} Certified robust radius $\delta'^{*}_{\mathrm{cert}}$, NNBC $\mathcal{B}_\varphi^*$, violation probability $\epsilon$;\\
\hrule
\vspace{0.5em}
\textbf{Step 1 - Data Generation }
\vspace{0.5em}
\hrule
\vspace{0.5em}
\begin{algorithmic}[1]
\FOR{$i = 1$ to $N$}
    \STATE Initialize ${\theta}_i{\mbox{\small$(0)$}} \in \Theta _0$ and add to $\mathcal{I}$;
    \FOR{$j = 1$ to $t_\infty$}
        \STATE Train $h_{\theta}$ on $\mathcal{D}_{\text{train}}$ to obtain ${\theta}_i(j)$;
    \ENDFOR 
    \STATE Sample poisoning level $\delta'_i$;
    \STATE Create $\rho'$-ratio poisoned dataset $\mathcal{D}_{\text{test}}^{\Delta'_i}$ with $\|\Delta'_i\|_p\!=\! \delta'_i$;
    \STATE Evaluate test accuracy $\mathcal{G}(\mbox{\small$\theta_i(t_\infty)$})$ on $\mathcal{D}_{\text{test}}^{\Delta'_i}$ as in definition~\ref{def:deg};
    \IF{$\mathcal{G}(\mbox{\small$\theta_i(t_\infty)$}) \leq \alpha$}
        \STATE Add ${\theta}_i{\mbox{\small$(t_\infty)$}}$ to $\mathcal{S}$;
    \ELSE
        \STATE Add ${\theta}_i{\mbox{\small$(t_\infty)$}}$ to $\mathcal{U}$;
    \ENDIF
\ENDFOR
\STATE Set $\vartheta\gets \mathcal{I} \cup \mathcal{U} \cup \mathcal{S}$;
\STATE Compute $\delta'_{\mathrm{emp}} \gets \max \left\{ \delta'_i \mid \mathcal{G}({\theta}_i{\mbox{\small$(t_\infty)$}}) \leq \alpha \right\}$;
\RETURN $\mathcal{I}, \mathcal{U}, \vartheta, \delta'_{\mathrm{emp}}$
\end{algorithmic}

\vspace{2em}
\hrule
\vspace{0.5em}
\textbf{Step 2 - Certification Framework}
\vspace{0.5em}
\hrule
\vspace{0.5em}

\begin{algorithmic}[1]
\STATE Generate $N$ trajectories to form $\mathcal{I}, \mathcal{U}, \vartheta$ along with corresponding empirical robust radius $\delta_{\mathrm{emp}}$ and generate $\hat{N}$ i.i.d. samples to form $\mathcal{Z}_1, \mathcal{Z}_2, \vartheta'$;
\STATE Initialize $\delta'_{\mathrm{cert}} \gets \delta'_{\mathrm{emp}}$, NNBC $\mathcal{B}_\varphi$, and counter $k \gets 0$;
\WHILE{$\delta'_{\mathrm{cert}} > 0$}
    \WHILE{$\mathcal{L} \neq 0$ \textbf{and} $k < T$}
        \STATE Train $\mathcal{B}_\varphi$ using $\mathcal{I}, \mathcal{U}, \vartheta$ with loss $\mathcal{L}$ as in \eqref{Loss_NNBC};
        \STATE Update $\mathcal{L}$ and increment $k \gets k + 1$;
        \IF{$\mathcal{L} \neq 0$ \textbf{and} $k\geq T$}
            \STATE Decrease radius: $\delta'_{\mathrm{cert}} \gets \delta'_{\mathrm{cert}} - d_\delta$;   \textbf{break}
        \ELSIF{$\mathcal{L} = 0$}
            \STATE Solve SCP as in \eqref{eq:scp} using $\mathcal{Z}_1, \mathcal{Z}_2, \vartheta'$ to obtain margin $\eta_s^*$;
            \IF{$\eta_s^* > 0$}
                \STATE Decrease radius: $\delta'_{\mathrm{cert}} \gets \delta'_{\mathrm{cert}} - d_\delta$;
                \STATE (Optional: Increase $N$ );
            \ELSE
                \STATE $\delta'^{*}_{\mathrm{cert}} \gets \delta'_{\mathrm{cert}}$;
                \STATE $\mathcal{B}_\varphi^* \gets \mathcal{B}_\varphi$;
                \STATE Compute $\epsilon$ from $\hat{N} = \left\lceil \frac{\ln(\beta)}{\ln(1 - \epsilon)} \right\rceil$;
                \STATE \textbf{break}
            \ENDIF
        \ENDIF
    \ENDWHILE
\ENDWHILE
\RETURN $\mathcal{B}_\varphi^*, \delta'^{*}_{\mathrm{cert}}, \epsilon$
\end{algorithmic}
\label{alg2}
\end{algorithm}

\subsection{ADDITIONAL EXPERIMENTS} \label{app:sec_exp}
\subsubsection{Results}
\begin{figure*}[!htb]
\centering

\begin{minipage}{0.185\linewidth}
    \centering
    \includegraphics[width=\linewidth]{Fig/1.png}\\[-2pt]
    \tiny (a) MNIST, CNN, PGD Train-Time
\end{minipage}\hfill
\begin{minipage}{0.185\linewidth}
    \centering
    \includegraphics[width=\linewidth]{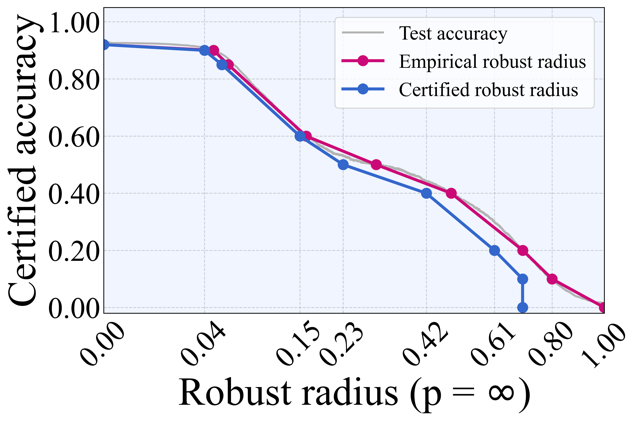}\\[-2pt]
    \tiny (b) SVHN, CNN, PGD Train-Time
\end{minipage}\hfill
\begin{minipage}{0.185\linewidth}
    \centering
    \includegraphics[width=\linewidth]{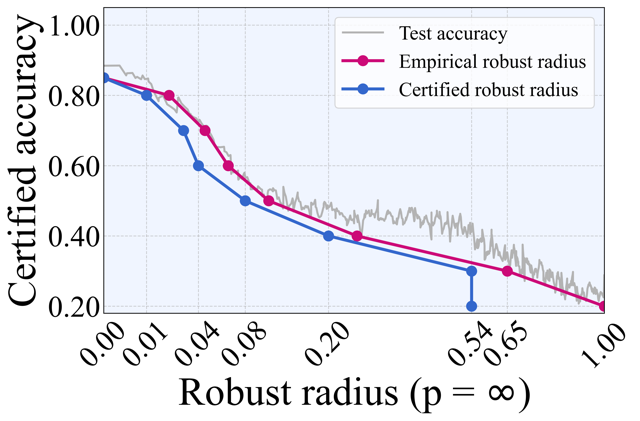}\\[-2pt]
    \tiny (c) CIFAR-10, CNN, PGD Train-Time
\end{minipage}\hfill
\begin{minipage}{0.185\linewidth}
    \centering
    \includegraphics[width=\linewidth]{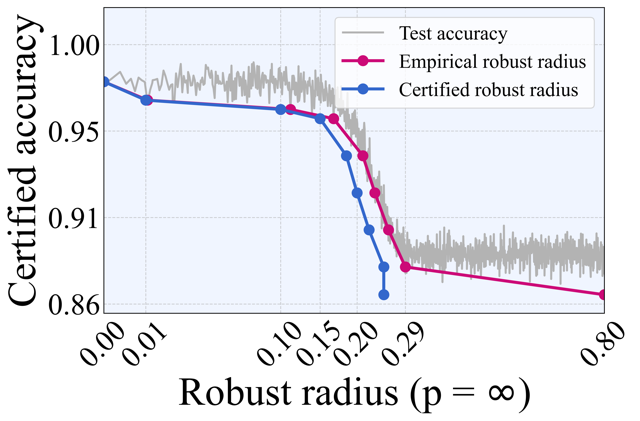}\\[-2pt]
    \tiny (d) MNIST, MLP, BPA Train-Time
\end{minipage}\hfill
\begin{minipage}{0.185\linewidth}
    \centering
    \includegraphics[width=\linewidth]{Fig/5.png}\\[-2pt]
    \tiny (e) SVHN, MLP, BDA Train-Time
\end{minipage}\\[0.7em]

\begin{minipage}{0.185\linewidth}
    \centering
    \includegraphics[width=\linewidth]{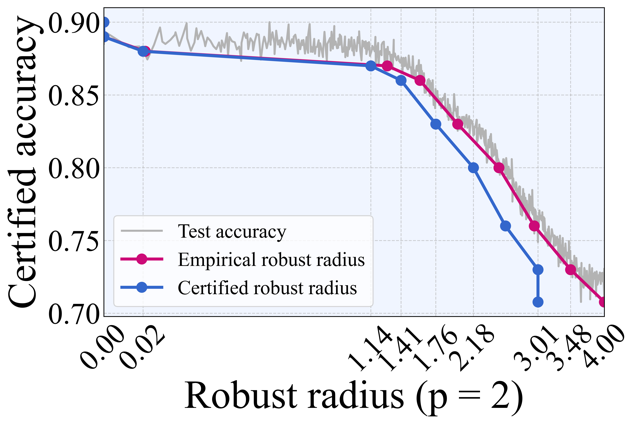}\\[-2pt]
    \tiny (f) SVHN, MLP, PGD Train-Time
\end{minipage}\hfill
\begin{minipage}{0.185\linewidth}
    \centering
    \includegraphics[width=\linewidth]{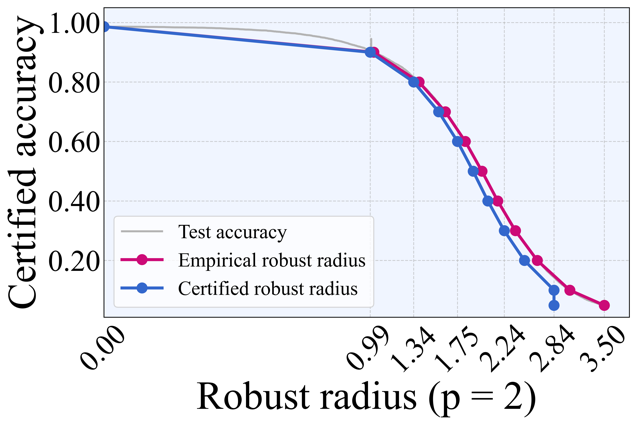}\\[-2pt]
    \tiny (g) MNIST, CNN, PGD Train-Time
\end{minipage}\hfill
\begin{minipage}{0.185\linewidth}
    \centering
    \includegraphics[width=\linewidth]{Fig/8.png}\\[-2pt]
    \tiny (h) MNIST, MLP, PGD Train-Time
\end{minipage}\hfill
\begin{minipage}{0.185\linewidth}
    \centering
    \includegraphics[width=\linewidth]{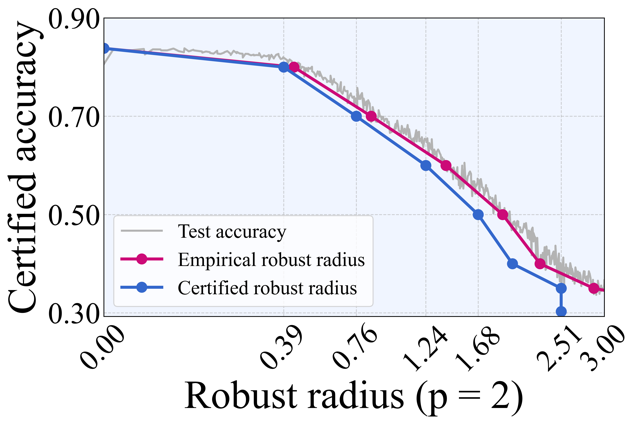}\\[-2pt]
    \tiny (i) SVHN, CNN, PGD Train-Time
\end{minipage}\hfill
\begin{minipage}{0.185\linewidth}
    \centering
    \includegraphics[width=\linewidth]{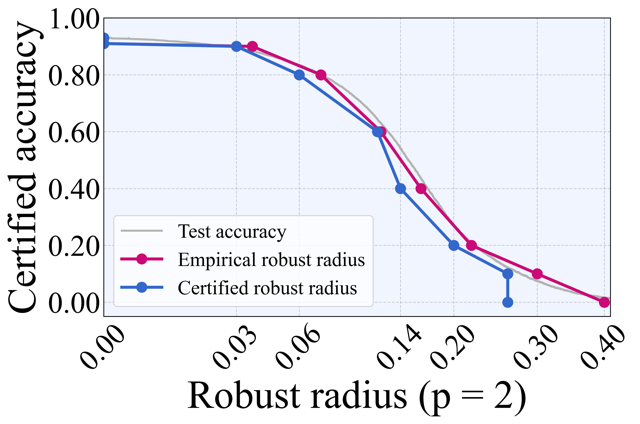}\\[-2pt]
    \tiny (j) SVHN, CNN, BDA Test-Time
\end{minipage}\\[0.7em]

\begin{minipage}{0.185\linewidth}
    \centering
    \includegraphics[width=\linewidth]{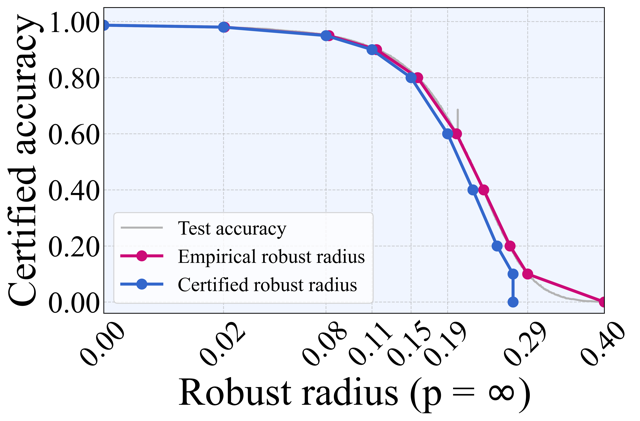}\\[-2pt]
    \tiny (k) MNIST, CNN, BDA Test-Time
\end{minipage}\hfill
\begin{minipage}{0.185\linewidth}
    \centering
    \includegraphics[width=\linewidth]{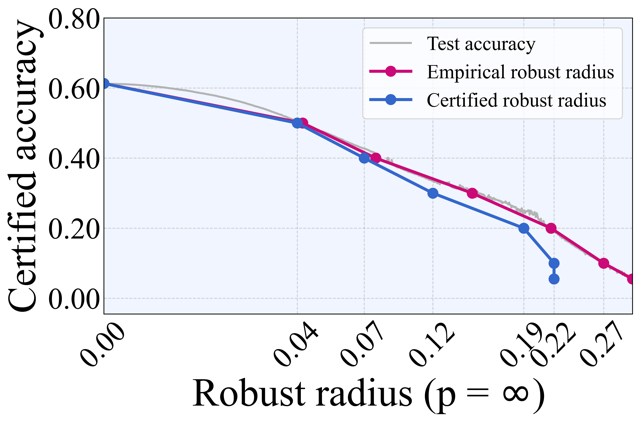}\\[-2pt]
    \tiny (l) CIFAR-10, CNN, BDA Test-Time
\end{minipage}\hfill
\begin{minipage}{0.185\linewidth}
    \centering
    \includegraphics[width=\linewidth]{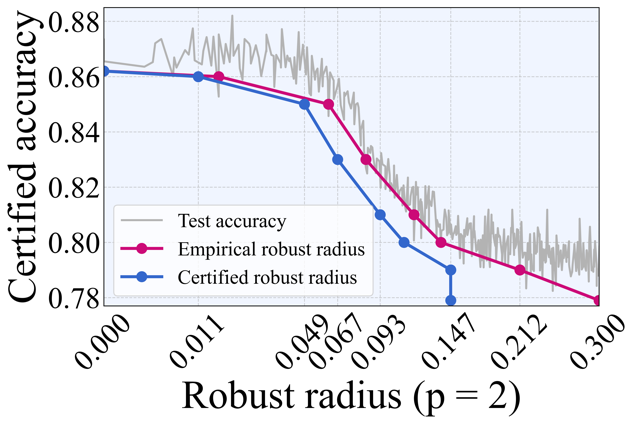}\\[-2pt]
    \tiny (m) SVHN, CNN, AA Test-Time
\end{minipage}\hfill
\begin{minipage}{0.185\linewidth}
    \centering
    \includegraphics[width=\linewidth]{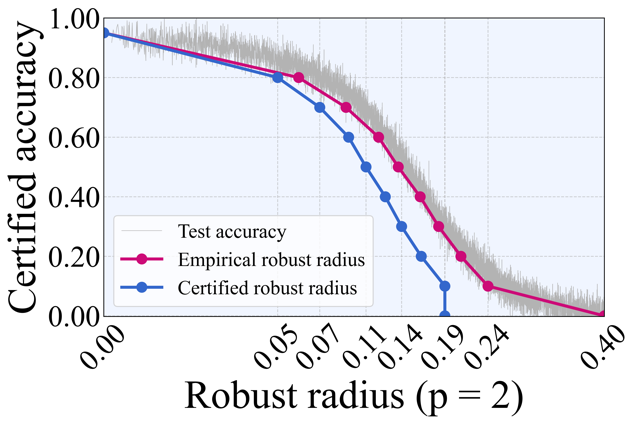}\\[-2pt]
    \tiny (n) MNIST, MLP, AA Test-Time
\end{minipage}\hfill
\begin{minipage}{0.185\linewidth}
    \centering
    \includegraphics[width=\linewidth]{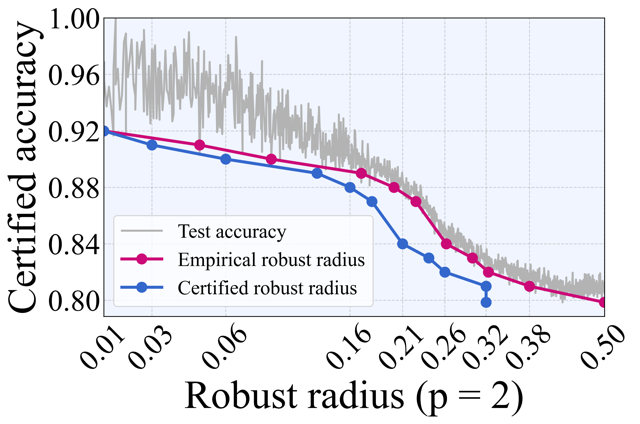}\\[-2pt]
    \tiny (o) SVHN, MLP, AA Test-Time
\end{minipage}\\[0.7em]

\begin{minipage}{0.185\linewidth}
    \centering
    \includegraphics[width=\linewidth]{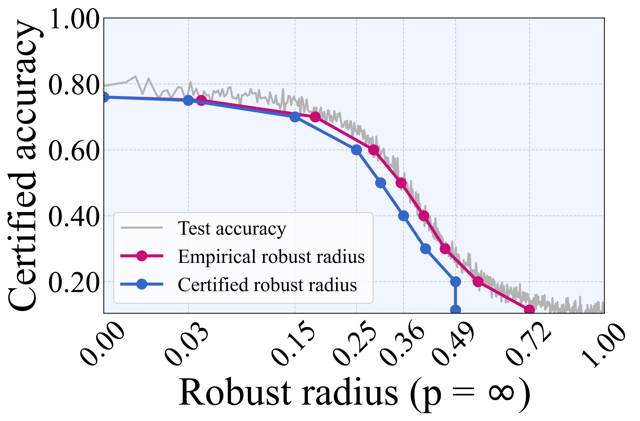}\\[-2pt]
    \tiny (p) SVHN, ResNet, BPA Train-Time
\end{minipage}\hfill
\begin{minipage}{0.185\linewidth}
    \centering
    \includegraphics[width=\linewidth]{Fig/17.png}\\[-2pt]
    \tiny (q) CIFAR-10, ResNet, BPA Train-Time
\end{minipage}\hfill
\begin{minipage}{0.185\linewidth}
    \centering
    \includegraphics[width=\linewidth]{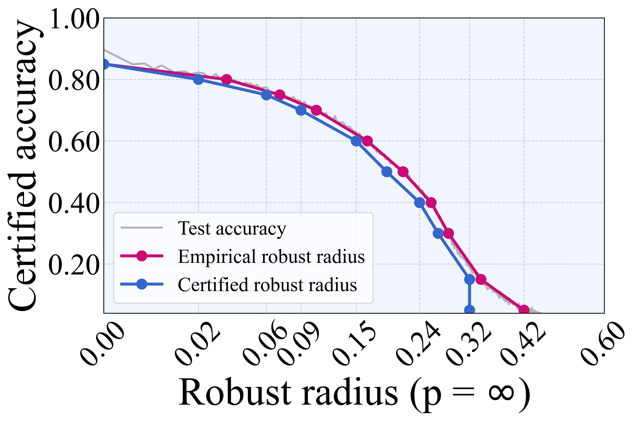}\\[-2pt]
    \tiny (r) CIFAR-10, ResNet, BDA Train-Time
\end{minipage}\hfill
\begin{minipage}{0.185\linewidth}
    \centering
    \includegraphics[width=\linewidth]{Fig/19.png}\\[-2pt]
    \tiny (s) SVHN, ResNet, PGD Train-Time
\end{minipage}\hfill
\begin{minipage}{0.185\linewidth}
    \centering
    \includegraphics[width=\linewidth]{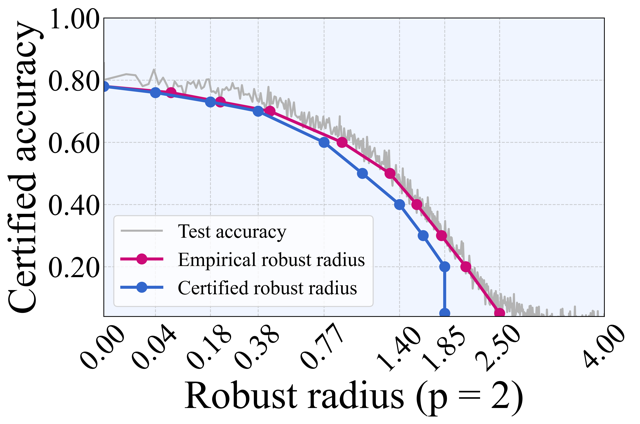}\\[-2pt]
    \tiny (t) CIFAR-10, ResNet, AA Test-Time
\end{minipage}
    \caption{
        Certified accuracy versus perturbation magnitude $\delta$ under different poisoning scenarios and datasets. Each subplot shows the test accuracy $g$, empirical robust radius $\delta_\mathrm{emp}$, and certified robust radius $\delta^*_\mathrm{cert}$ under the proposed framework. The confidence level is fixed at $99.99\%$. Violation probabilities are: $\epsilon = $ (a) $0.005$, (b) $0.011$, (c) $0.045$, (d) $0.003$, (e) $0.006$, (f) $0.006$, (g) $0.006$, (h) $0.004$, (i) $0.011$, (j) $0.015$, (k) $0.006$, (l) $0.045$, (m) $0.009$, (n) $0.003$, (o) $0.004$, (p) $0.009$, (q) $0.015$, (r) $0.015$, (s) $0.015$, (t) $0.022$. 
    }
\label{fig:All_certified-accuracy-grid}
\end{figure*}

Figure~\ref{fig:All_certified-accuracy-grid} reports certified accuracy $g_\mathrm{p}^*$ versus perturbation radius $\delta$ across attacks, models, datasets, and both train- and test-time settings. In most cases, the certified robust radius $\delta^{*}_{\mathrm{cert}}$ remains close to the empirical radius $\delta^{*}_{\mathrm{emp}}$, especially when test accuracy degrades smoothly, indicating tight certificates. A quantitative comparison with RAB under clean-label backdoor attacks (Table~\ref{tab:framevsRAB}) shows comparable certified radii, while our method more closely follows empirical robustness.
The NNBC $\mathcal{B}_\varphi$ is trained with Adam using the multi-term loss in~\eqref{Loss_NNBC}. To balance the penalties across constraint sets, we normalize the weights by set size:
$c_{\mathcal{I}}\!=\!\tfrac{1}{N_{\mathcal{I}}}$,\;
$c_{\mathcal{U}}\!=\!\tfrac{1}{N_{\mathcal{U}}}$,\;
$c_{\mathcal{Z}}\!=\!\tfrac{1}{N_{\mathcal{Z}}}$,
where $N_{\mathcal{I}}, N_{\mathcal{U}}, N_{\mathcal{Z}}$ are the cardinalities of the initial, unsafe, and feasible sublevel sets, respectively. This normalization compensates for variations induced by the choice of the gap threshold $\alpha$.

\subsubsection{Attack models} 
Table~\ref{tab:attack_models} summarizes the train-time poisoning and test-time evasion attacks used in our experiments, including their category, adversarial access, and objective/threat model under the generic $\ell_p$-bounded setting of Definitions~\ref{def:poisoning}--\ref{def:test_evasion}.
The \emph{Category} column distinguishes gradient-based attacks, backdoor attacks, clean-label poisoning, decision-boundary attacks, and score-based black-box attacks. The \emph{Access} column specifies whether the adversary is white-box, black-box, or data-level. The \emph{Objective / threat model} column indicates which samples are perturbed, whether the attack is targeted or untargeted, and how the perturbation budget is enforced.
In all cases, $\delta$ denotes the per-sample $\ell_p$ norm bound on the input perturbations, while train-time poisoning attacks additionally satisfy a fixed poison ratio $\rho$ controlling the fraction of modified samples. The table serves as a concise link between the concrete attacks used in the experiments and the abstract threat model of our certification framework.
\begin{table*}[!htb]
\centering
{
\caption{Train-time (poisoning) and test-time (evasion) attacks used in additional experiments. All attacks are implemented under an $\ell_p$-bounded threat model with maximum perturbation magnitude $\delta$.}
\label{tab:attack_models}
\centering
\renewcommand{\arraystretch}{1}
\scalebox{0.65}{
\setlength{\tabcolsep}{5pt}
\begin{tabular}{m{0.8cm}|m{2.5cm}|m{3.6cm}|m{1.4cm}|m{16cm}} 

\toprule
\textbf{Time} & \textbf{Attack} & \textbf{Category} & \textbf{Access} & \textbf{Objective / threat model} \\
\midrule

\multirow{6}{*}{\textbf{Train}}
& PGD 
& Gradient-based poisoning
& White-box
& Untargeted train-time attack: Perturb training inputs $u_i \mapsto u_i+\delta_i$ with $\|\delta_i\|_p \le \delta$ via projected gradient ascent to increase training loss and degrade test accuracy. 
\\[0.35em]
\cmidrule(lr){2-5} 
& Backdoor Attack
& Backdoor poisoning
& Data-level
& Untargeted train-time attack: Add a fixed trigger perturbation $\delta_{\mathrm{bd}}$ with $\|\delta_{\mathrm{bd}}\|_p \le \delta$ to a fraction of training inputs $u_i \mapsto u_i + \delta_{\mathrm{bd}}$ so that the trigger induces a targeted backdoor behavior. 
\\[0.35em]
\cmidrule(lr){2-5} 
& Bullseye Polytope
& Data geometry-based poisoning 
& White-box
& Targeted train-time attack: Perturb training inputs $u_i \mapsto u_i+\delta_i$ with $\|\delta_i\|_p \le \delta$ so that the target lies close to the center of the poison samples' polytope 
while keeping labels unchanged. 
\\

\midrule
\multirow{9}{*}{\textbf{Test}}
& Auto PGD
& Gradient-based evasion
& White-box
& Untargeted test-time attack: perturb each test input $u_i' \mapsto u_i' + \delta_i'$ with $\|\delta_i'\|_p \le \delta$ via projected gradient ascent using adaptive step size to increase loss and cause misclassification. 
\\[0.35em]
\cmidrule(lr){2-5} 
& FAB
& Decision-boundary
& White-box
& Untargeted decision-boundary attack: search for an (approximately) minimal-norm $\delta_i'$ with $\|\delta_i'\|_p \le \delta$ such that prediction of $u_i' + \delta_i'$ is not the same as its original label $y_i'$. 
\\[0.35em]
\cmidrule(lr){2-5} 
& Square Attack
& Score-based evasion
& Black-box
& Untargeted black-box attack: apply randomized square-shaped updates to $u_i'$ within $\|\delta_i'\|_p \le \delta$ to increase loss using only model scores. 
\\[0.35em]
\cmidrule(lr){2-5} 
& Backdoor Attack
& Backdoor evasion
& Data-level
& Targeted test-time attack: add the learned trigger perturbation $\delta_{\mathrm{bd}}'$ with $\|\delta_{\mathrm{bd}}'\|_p \le \delta$ to inputs $u_i' \mapsto u_i' + \delta_{\mathrm{bd}}'$ to activate the backdoor mapping. \\
\bottomrule
\end{tabular}
}}
\end{table*}

\subsubsection{Configurations} All experiments were implemented in PyTorch (Python~3.11) and run on two environments:
(A) a MacBook Pro with Apple M3 Pro (12-core CPU), 36\,GB RAM, macOS Sonoma 14.4; 
(B) 4$\times$~NVIDIA H100 GPUs with a 32-core CPU, and 128\,GB RAM.
Hardware configurations are denoted abstractly as \textbf{A} and \textbf{B} in the tables. Full experimental settings appear in Table~\ref{tab:configs}, with corresponding figures in the last column; model architectures are listed in Table~\ref{tab:model-specs}.

\begin{table*}[!t]
\caption{Experimental configurations across datasets, attacks, optimizers, models, and NNBC settings.}
\label{tab:configs}
\centering
\renewcommand{\arraystretch}{1}
\scalebox{0.65}{
\setlength{\tabcolsep}{8pt}
\begin{tabular}{cccccccccccccccccc}
\toprule
\multirow{3}{*}{\textbf{Cfg.}} & \multirow{3}{*}{\textbf{Dataset}} & 
\multicolumn{4}{c}{\textbf{Attack}} & 
\multicolumn{2}{c}{\textbf{ML Setup}} & 
\multicolumn{6}{c}{\textbf{Certificate}} &
\multicolumn{2}{c}{\textbf{Execution Setup}} &
\multirow{3}{*}{\textbf{Fig./Tab.}} \\
\cmidrule(lr){3-6} \cmidrule(lr){7-8} \cmidrule(lr){9-14} \cmidrule(lr){15-16}
& & \multirow{2}{*}{\textbf{Type}} 
& \multirow{2}{*}{\textbf{$\boldsymbol{p}$-norm}} 
& \multirow{2}{*}{\textbf{$\boldsymbol{\rho}$}} 
& \multirow{2}{*}{\textbf{Step}}
& \multirow{2}{*}{\textbf{Model}} 
& \multirow{2}{*}{\textbf{Optimizer ($\boldsymbol{l_r}$)}}
& \multirow{2}{*}{\textbf{Type}} 
& \multicolumn{3}{c}{\textbf{NNBC}} 
& \multicolumn{2}{c}{\textbf{PAC}}
& \multirow{2}{*}{\textbf{HW}} 
& \multirow{2}{*}{\textbf{Runtime}} 
& \\
\cmidrule(lr){10-12} \cmidrule(lr){13-14}
& & & & & & & &
& \textbf{$\boldsymbol{N}$} 
& \textbf{Layer} 
& \textbf{Optimizer ($\boldsymbol{l_r}$)} 
& \textbf{$\boldsymbol{\hat{N}}$} 
& \textbf{$\boldsymbol{\epsilon}$}
& & (min)& \\
\midrule
1  & MNIST   & PGD & $\infty$ & 1 & 40 & CNN & SGD (0.01)   & Train-Time & $4000$ & 5 & Adam (0.001) & 1800  & 0.005 & A & 53  & Fig.5a \\
2  & MNIST   & BPA & $\infty$ & 0.3 & 30& MLP & GD (0.10)           & Train-Time & $5000$ & 7 & Adam (0.001) & 2500 & 0.003 & A & 36  & Fig.5d \\
3  & MNIST   & PGD & 2        & 1 & 40 & CNN & Adam (0.001) & Train-Time & $3000$ & 5 & Adam (0.01)   & 1500  & 0.006 & A & 19  & Fig.5g \\
4  & MNIST   & PGD & $\infty$ & 1 & 40 & MLP & SGD (0.01)   & Train-Time & $4000$ & 5 & Adam (0.001) & 2000  & 0.004 & A & 31  & Fig.5h \\
5  & MNIST   & BDA & $\infty$ & 0.1 & 40 & CNN & SGD (0.01) & Test-Time  & $3000$ & 5 & Adam (0.001) & 1500  & 0.006 & A & 18  & Fig.5k \\
6  & MNIST   & AA  & 2        & 1 &100 & MLP & Adam (0.001) & Test-Time  & $4000$ & 7 & Adam (0.10)    & 2500 & 0.003 & A & 26  & Fig.5n \\\midrule
7  & SVHN    & PGD & $\infty$ & 0.9 & 40 & CNN & Adam (0.001) & Train-Time & $2000$ & 4 & Adam (0.001)   & 800  & 0.011 & A & 39  & Fig.5b \\
8  & SVHN    & BDA & $\infty$ & 0.1 & 30& MLP & GD (0.10)    & Train-Time & $4000$ & 5 & Adam (0.001) & 1500  & 0.006 & B &  13 & Fig.5e \\
9  & SVHN    & PGD & 2 & 0.9 & 30& MLP & SGD (0.01)    & Train-Time & $4000$ & 5 & Adam (0.001) & 1500  & 0.006 & B & 27  & Fig.5f \\
10 & SVHN    & PGD & 2        & 0.9 & 40 & CNN & SGD (0.01)   & Train-Time & $2000$ & 4 & Adam (0.001) & 800  & 0.011 & A & 73  & Fig.5i \\
11 & SVHN    & BDA & 2 & 0.9 & 40 & CNN & SGD (0.10)    & Test-Time  & $2500$ & 4 & Adam (0.001) & 600  & 0.015 & A & 25 & Fig.5j \\
12 & SVHN    & AA  & 2        & 0.8 &100 & CNN & SGD (0.01)   & Test-Time  & $3000$ & 4 & Adam (0.001) & 1000 & 0.009 & A & 81 & Fig.5m \\
13 & SVHN    & AA  & 2        & 0.8 &100 & MLP & GD (0.10)    & Test-Time  & $4000$ & 5 & Adam (0.001) & 2000 & 0.004 & A & 63 & Fig.5o \\
14 & SVHN    & BPA & $\infty$ & 0.2 & 30& ResNet & Adam (0.10)    & Train-Time & $4000$ & 4 & Adam (0.001) & 1000  & 0.009 & B & 72  & Fig.5p \\
15 & SVHN    & PGD & 2        & 0.9 & 40 & ResNet & Adam (0.01)   & Train-Time & $3000$ & 5 & Adam (0.001) & 700  & 0.015 & B & 84  & Fig.5s \\
\midrule
16 & CIFAR10 & PGD & $\infty$ & 0.8 & 40 & CNN & Adam (0.001) & Train-Time & $1500$ & 4 & Adam (0.001)   & 200  & 0.045 & A & 138  & Fig.5c \\
17 & CIFAR10 & BDA & $\infty$ & 0.1 & 40 & CNN & Adam (0.001)  & Test-Time  & $1500$ & 4 & Adam (0.001) & 200  & 0.045 & B & 66  & Fig.5l \\
18 & CIFAR10 & BPA & 2        & 0.2 & 30 & ResNet & Adam (0.01)   & Train-Time & $2000$ & 4 & SGD (0.001) & 600  &0.015 & B & 81  & Fig.5q \\
19 & CIFAR10 & BDA & $\infty$ & 0.2 & 40 & ResNet & Adam (0.01)  & Train-Time  & $3000$ & 4 & Adam (0.001) & 600  & 0.015 & B & 76  & Fig.5r \\
20 & CIFAR10 & AA & 2 & 0.3 & 80 & ResNet & SGD (0.01)   & Test-Time & $2000$ & 4 & Adam (0.001) & 400  & 0.022 & B & 98  & Fig.5t \\
\midrule
21 & CIFAR100 & PGD & 2 & 0.6 & 20 & ResNet & Adam (0.01)  & Train-Time  & $800$ & 5 & Adam (0.01)  & $200$  & 0.045 & B & 228  & Tab.1 \\
22 & CIFAR100 & BPA & $\infty$ & 0.2 & 60 & ResNet & Adam (0.01)  & Train-Time  & $1000$ & 4 & Adam (0.001)  & $200$  & 0.045 & B & 182  & Tab.1 \\
\bottomrule
\end{tabular}
}
\end{table*}

\begin{table*}[!htb]
\caption{Architectural specifications of models used across datasets.}
\label{tab:model-specs}
\centering
\renewcommand{\arraystretch}{0.95}
\scalebox{0.63}{ 
\setlength{\tabcolsep}{32pt}
\begin{tabular}{cccccc}
\toprule
\textbf{Dataset} & \textbf{Model} & \textbf{Conv Layers} & \textbf{Pooling} & \textbf{FC Layers} & \textbf{Params (M)} \\
\midrule
\multirow{2}{*}{MNIST} 
 & CNN      & 3 conv (32, 64, 128) & 3× MaxPool (2×2) & 3 FC (256, 128, 10) & $\sim$1.2M \\
 & MLP      & -- & -- & 4 FC (512, 256, 128, 10) & $\sim$0.6M 
 \\
\midrule
\multirow{3}{*}{SVHN} 
 & CNN      & 2 conv (64, 128) & 2× MaxPool (2×2) & 3 FC (256, 128, 10) & $\sim$1.5M \\
 & MLP      & -- & -- & 4 FC (1024, 512, 256, 10) & $\sim$3.2M \\
 & ResNet18 & 18 conv (standard) & Global AvgPool & 1 FC (10) & $\sim$11.18M \\
\midrule
\multirow{2}{*}{CIFAR-10} 
 & CNN      & 3 conv (64, 128, 256) & 3× MaxPool (2×2) & 3 FC (512, 256, 10) & $\sim$4.5M \\
 & ResNet18 & 18 conv (standard) & Global AvgPool & 1 FC (10) & $\sim$11.18M \\
\midrule
CIFAR-100
 & ResNet18 & 18 conv (standard) & Global AvgPool & 1 FC (100) & $\sim$11.23M \\
\bottomrule
\end{tabular}
}
\vspace{-0.4cm}
\end{table*}

\subsection{ADDITIONAL DISCUSSION}
\label{app:discussion}
\subsubsection{Fully agnostic framework}
Our framework models gradient-based training as a discrete-time dynamical system in parameter space and relies only on observed parameter trajectories and the terminal safety condition $\mathcal{G}(\mbox{\small$\theta$}) \le \alpha$. It does not require explicit knowledge of the attack strategy, trigger, poisoning ratio, architecture, loss landscape, or optimizer/scheduler, as these are absorbed into the realized update map $f$ and reflected through the empirical reachable set used to train and verify the BC $\mathcal{B}(\theta)$.
Under this abstraction, the same construction applies to both train-time and test-time feature-space poisoning across different model classes and training pipelines, without attack-specific tuning. This broad applicability extends to high-capacity models. However, the main cost of this generality is computational, since NNBC learning and scenario generation require multiple empirical training trajectories (see Table~\ref{tab:configs}).

\subsubsection{Empty sets and robust radius adjustment}
Safe/unsafe labels in Section~\ref{subsec:data_gen} are defined by a threshold $\alpha\in[0,1]$ applied to the test accuracy $g(\theta_i(t_\infty))$. If $\alpha$ exceeds the clean accuracy at $\delta=0$, then $\mathcal{S}=\emptyset$; if it is below the worst-case accuracy at $\delta=\delta_{\max}$, then $\mathcal{U}=\emptyset$.
These degenerate cases prevent computation of the empirical margin $\eta_s^*$ in the SCP. Thus, if $\mathcal{S}=\emptyset$, we set $\delta_{\mathrm{cert}}$ (resp.\ $\delta_{\mathrm{cert}}'$) to zero. If $\mathcal{U}=\emptyset$, we increase $\alpha$ until both sets are non-empty and verification succeeds, and take the smallest such value as the effective certified $\alpha$.
This also explains why, in plots such as Figure~\ref{fig:certified-accuracy-grid}, the certified radius may remain unchanged even for lower $\alpha$. In such cases, certification at the lower threshold fails due to infeasibility or insufficient data, so we conservatively retain the last valid radius.

\subsubsection{Effect of model strength on certificate tightness}
We sometimes observe tighter certificates, i.e., $\delta^*_{\mathrm{cert}}$ closer to $\delta^*_{\mathrm{emp}}$, in harder settings such as CIFAR-100 with ResNet. This is not due to dataset difficulty itself, but to the stronger architectures these settings require. Such models usually achieve higher clean accuracy and, more importantly, smoother accuracy degradation as the perturbation radius $\delta$ increases. Since the barrier is learned from this accuracy--radius profile, smoother trajectories make the safe set easier to approximate and can yield tighter certified radii. The trade-off is computational: harder settings are more expensive, so $\hat{N}$ is often smaller, which weakens the PAC bound and increases the certified violation probability $\epsilon$.

\subsubsection{On seemingly extreme poisoning ratios} 
In several experiments we deliberately consider very large corruption ratios (e.g., $0.5$–$1$), even though such levels are rare in practice. This is intentional and reflects what our framework actually certifies: a bound on the per-sample perturbation magnitude (the robust radius), without assuming any fixed fraction of corrupted points. The corruption fraction is treated as an unknown in $[0,1]$ and does not appear in the certificate itself. Using such extreme ratios allows us  (1)
to stress-test the method in challenging regimes and highlight
that our guarantees are independent of the corrupted fraction,
and (2) to contrast with prior work that fixes this ratio and
certifies how many samples can be poisoned.


\subsubsection{\texorpdfstring{Effect of random Initialization on $\delta_{\mathrm{emp}}$}{Effect of random Initialization on delta emp}}\label{para:random_emp}
The empirical radius $\delta_{\mathrm{emp}}$ is computed from finitely many randomly initialized training trajectories and thus depends on the initialization scheme. For reproducibility, we fix the random seed when reporting certificates.
Without a fixed seed, $\delta_{\mathrm{emp}}$ may vary across reruns. Table~\ref{tab:rerun_delta_emp} reports its values over five independent reruns for three representative configurations, together with the corresponding mean and standard deviation, showing that this variation is typically moderate.
Importantly, $\delta_{\mathrm{emp}}$ is not the final robustness claim. It only initializes the search over admissible budgets for barrier-certificate synthesis. The final certified radius is reported only after the candidate certificate passes the subsequent scenario-based verification stage. Thus, reruns mainly affect the stability of this empirical upper bound and the associated datasets, while the final certification decision is made only at verification.

\begin{table}[!t]
\centering
{
\caption{Empirical robust radius $\delta_{\mathrm{emp}}$ across five independent reruns for three representative configurations.}
\label{tab:rerun_delta_emp}
\renewcommand{\arraystretch}{1.1}
\setlength{\tabcolsep}{5pt}
\scalebox{0.75}{
\begin{tabular}{cc|ccccc|c}
\toprule
\textbf{Cfg.} & $\boldsymbol{g_{\mathrm{p}}^*}$ & \textbf{Run~1} & \textbf{Run~2} & \textbf{Run~3} & \textbf{Run~4} & \textbf{Run~5} & \textbf{Mean $\pm$ Std} \\
\midrule
$1$  & $0.9$  & $0.148$ & $0.146$ & $0.149$ & $0.148$ & $0.147$ & $0.1476 \pm 0.0011$ \\
\midrule
$8$  & $0.8$  & $0.252$ & $0.255$ & $0.251$ & $0.249$ & $0.257$ & $0.2528 \pm 0.0030$ \\
\midrule
$15$ & $0.75$ & $1.215$ & $1.228$ & $1.203$ & $1.217$ & $1.221$ & $1.2168 \pm 0.0092$ \\
\bottomrule
\end{tabular}
}
}
\end{table}

\subsubsection{Synthesis vs.\ verification}\label{syn_vs_ver}
In our framework, synthesis is purely empirical: it learns a candidate NNBC from sampled safe and unsafe trajectories and uses the empirical robust radius $\delta_{\mathrm{emp}}$ (resp. $\delta_{\mathrm{emp}}'$ for test-time) only as an initial candidate budget. Since $\delta_{\mathrm{emp}}$ is computed from finitely many sampled trajectories, it is not itself a certified claim.
The certified radii $\delta_{\mathrm{cert}}$ and $\delta_{\mathrm{cert}}'$ are determined only in the subsequent verification stage, where the candidate certificate is tested on fresh, independent trajectories through the SCP and its probabilistic guarantee. Hence, $\delta_{\mathrm{cert}} \!\le\! \delta_{\mathrm{emp}}$ and $\delta_{\mathrm{cert}}' \!\le\! \delta_{\mathrm{emp}}'$, with equality generally prevented by the conservativeness of finite-sample verification.
This distinction is essential: the synthesis loss is only a heuristic objective for separating empirically safe and unsafe trajectories, whereas the verification stage establishes the actual robustness claim by checking the barrier conditions on independent samples with explicit confidence parameters. The gap between empirical and certified radii depends on NNBC approximation quality, trajectory coverage, and the conservativeness of the scenario-based bound, and is not quantified explicitly in this work. Empirically, more regular parameter trajectories tend to produce tighter certificates.

\subsubsection{Role of RCP, CCP, SCP, and PAC.}\label{para:rcp/scp}
Direct certification over all admissible poisoning realizations is intractable, since it leads to infinitely many constraints over continuous uncertainty sets. We therefore use the standard relaxation chain RCP~$\Rightarrow$~CCP~$\Rightarrow$~SCP, where the RCP encodes the exact robustness requirement, the CCP allows violation probability at most $\epsilon$, and the SCP replaces the chance constraints with finitely many i.i.d.\ scenarios.
Problems~\eqref{eq:rcp}--\eqref{eq:scp} should be interpreted as margin-minimization problems rather than binary feasibility tests. If the RCP is feasible, then the CCP is also feasible. Moreover, the SCP is always well-posed as an optimization problem, since the sampled constraints are finite and $\eta_s\in\mathbb{R}$ is free. Thus, certification is determined by the sign of the optimal margin, namely whether $\eta_r^*\leq 0$ or, in the scenario-based setting, $\eta_s^*\leq 0$.
Under~\eqref{N_hat}, scenario theory~\cite{calafiore2006scenario,campi2008exact} guarantees that, with confidence at least $1-\beta$, the SCP solution is $\epsilon$-feasible for the CCP, meaning that the barrier conditions are violated with probability at most $\epsilon$. This yields a finite-sample probabilistic robustness certificate without additional structural assumptions on the training dynamics.

\subsubsection{Generality and future work}
Our framework models gradient-based training as a discrete-time dynamical system in parameter space and relies only on sampled parameter trajectories used to train and verify the BC $\mathcal{B}_\varphi$. It does not require white-box knowledge of the attack strategy, trigger, poisoning ratio, architecture, loss landscape, or optimizer, as these are absorbed into the realized update map $f$.
While we certify only feature-space perturbations, the framework remains general in that it covers both train-time and test-time poisoning, supports arbitrary model architectures and gradient-based optimizers such as (S)GD or Adam, and does not require knowledge of the attack strategy or corruption ratio. Extending the framework beyond ${\ell}_p$ perturbations is a natural direction for future work.

\bibliographystyle{IEEEtran}
\bibliography{CSYS}

\end{document}